\documentclass[10pt,journal,compsoc]{IEEEtran}

\usepackage{amsmath,amssymb} 
\usepackage{color}
\usepackage{booktabs}
\usepackage{multirow}
\usepackage{graphicx}
\usepackage{subfigure}
\usepackage{algorithm}
\usepackage{sidecap}
\usepackage[noend]{algpseudocode}
\usepackage[table]{xcolor}
\usepackage{bm}
\usepackage{enumitem}

\newcommand{\etal}{\textit{et al}.}

\newcommand{\STAB}[1]{\begin{tabular}{@{}c@{}}#1\end{tabular}}

\makeatother
\usepackage[pagebackref=false,breaklinks=true,letterpaper=true,colorlinks,bookmarks=false]{hyperref}

\newcommand\Tstrut{\rule{0pt}{2.2ex}}  
\ifCLASSOPTIONcompsoc
  \usepackage[nocompress]{cite}
\else
  \usepackage{cite}
\fi
\ifCLASSINFOpdf
\else
\fi

\begin{document}
\definecolor{task1}{RGB}{225, 243, 240}
\definecolor{task2}{RGB}{230, 242, 255}
\newcolumntype{a}{>{\columncolor[RGB]{225, 243, 240}}l}
\newcolumntype{b}{>{\columncolor[RGB]{230, 242, 255}}l}
\newcolumntype{g}{>{\columncolor[RGB]{249, 248, 239}}l}
\newcolumntype{d}{>{\columncolor[RGB]{249, 239, 239}}l}
\newcolumntype{e}{>{\columncolor[RGB]{234, 245, 246}}l}
\newcolumntype{f}{>{\columncolor[RGB]{249, 239, 249}}l}
\newcolumntype{k}{>{\columncolor[RGB]{255, 239, 213}}c}

\title{\vspace{-10pt}Incremental Object Detection via Meta-Learning}

\author{K J Joseph, 
Jathushan Rajasegaran, 
Salman Khan, Fahad Shahbaz Khan, Vineeth N Balasubramanian
\IEEEcompsocitemizethanks{\IEEEcompsocthanksitem K. J. Joseph and V. N. Balasubramanian are with the Department
of Computer Science and Engineering, Indian Institute of Technology Hyderabad,
India.
E-mail: cs17m18p100001@iith.ac.in
\IEEEcompsocthanksitem  J. Rajasegaran is with University of Calfornia, Berkeley, USA. 
\IEEEcompsocthanksitem 
S. H. Khan is with MBZ University of AI, Abu Dhabi, UAE and Australian National University, Canberra, Australia.
\IEEEcompsocthanksitem F. S. Khan is with MBZ University of AI, Abu Dhabi, UAE and CVL, Linköping University, Sweden.}
\thanks{Manuscript received 31 July 2020; revised 8 Mar. 2021; accepted 11 Oct. 2021.} 
}

\markboth{IEEE TRANSACTIONS ON PATTERN ANALYSIS AND MACHINE INTELLIGENCE, DOI 10.1109/TPAMI.2021.3124133}%
{~}

\IEEEtitleabstractindextext{%
\begin{abstract}
In a real-world setting, object instances from new classes can be continuously encountered by object detectors. When existing object detectors are applied to such scenarios, their performance on old classes deteriorates significantly. A few efforts have been reported to address this limitation, all of which apply variants of knowledge distillation to avoid catastrophic forgetting. We note that although distillation helps to retain previous learning, it obstructs fast adaptability to new tasks, which is a critical requirement for incremental learning. 
In this pursuit, we propose a meta-learning approach that learns to reshape model gradients, such that information across incremental tasks is optimally shared. This ensures a seamless information transfer via a meta-learned gradient preconditioning that minimizes forgetting and maximizes knowledge transfer. In comparison to existing meta-learning methods, our approach is task-agnostic, allows incremental addition of new-classes and scales to high-capacity
models for object detection. We evaluate our approach on a variety of incremental learning settings defined on PASCAL-VOC and MS COCO datasets, where our approach performs favourably well against state-of-the-art methods. Code and trained models: \url{https://github.com/JosephKJ/iOD}
\vspace{-5pt}
\end{abstract}

\begin{IEEEkeywords}
Object Detection, Incremental Learning, Deep Neural Networks, Meta-learning, Gradient preconditioning.
\end{IEEEkeywords} }

\maketitle
\IEEEdisplaynontitleabstractindextext

\IEEEpeerreviewmaketitle
\section{Introduction} \label{sec:introduction}
Deep learning has brought about remarkable improvements on numerous vision tasks, including object detection \cite{lin2017focal,ren2015faster,redmon2016you}. Most existing detection models make an inherent assumption that examples of all the object classes are available during the training phase. In reality, new classes of interest can be encountered on the go, due to the dynamic nature of the real-world. This makes the existing methods brittle in an incremental learning setting, wherein they tend to forget old task information when trained on a new task \cite{mccloskey1989catastrophic}.  


In this work, we study the \emph{class-incremental} object detection problem, where new classes are sequentially introduced to the detector. An intelligent learner must not forget previously learned classes, while learning to detect new object categories. To this end, knowledge distillation \cite{hinton2015distilling} has been utilized as a de facto solution \cite{shmelkov2017incremental,li2019rilod,hao2019end,chen2019new}.  While learning a new set of classes, distillation based methods ensure that the classification logits and the regression targets of the previous classes, are not altered significantly from the earlier state of the model. As a side effect, distillation enforces intransigence in the training procedure, making it hard to learn novel classes. An essential characteristic for incremental object detectors is to have optimal plasticity, which aids in quick adaptability to new classes without losing grasp of previously acquired knowledge.

Learning to learn for quick adaptability forms the basis of current meta-learning methods \cite{finn2017model,nichol2018first,snell2017prototypical}. These methods have generally been successful in few-shot learning settings. Directly adopting such methodologies to incremental object detection is challenging due to the following reasons: (a) A meta-learner is explicitly trained with a \emph{fixed} number of classes  ($N$-way classification) and does not generalize to an incremental setting. 
(b) Each task in the meta-train and meta-test stages is carefully designed to avoid task-overfitting \cite{yin2019meta}. This is prohibitive in an object detection setting as each image can possibly have multi-class instances. (c) The meta-learners require knowledge about the end-task for fine-tuning, while in the incremental setting a test sample can belong to any of the classes observed so-far, implicitly demanding task-agnostic inference. (d) Network architectures meta-learned in a traditional setting are in the order of a few convolutional layers. This comes in stark contrast to an object detector which involves multiple sub-networks for generating backbone features,  object proposals and final classification and localization outputs.


\begin{figure*}[t]
\centering
\subfigure[Learning $\mathcal{T}_1$]{\includegraphics[width=0.32\linewidth]{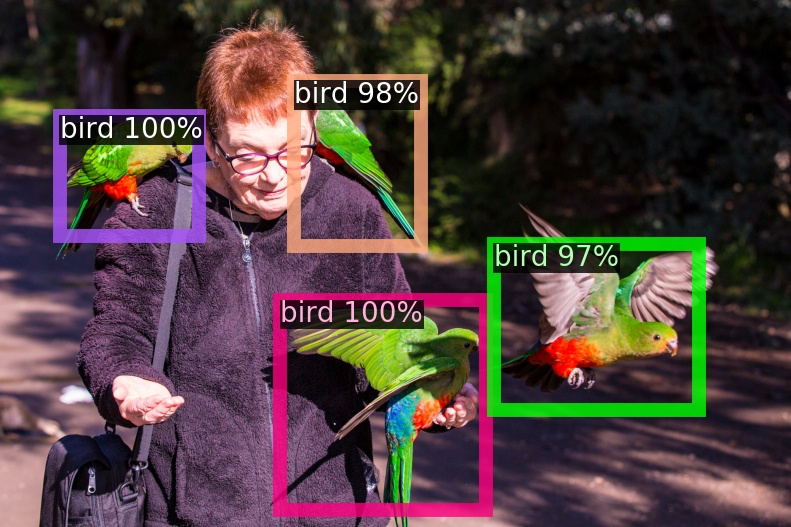}}
~\subfigure[Learning $\mathcal{T}_2$]{\includegraphics[width=0.32\linewidth]{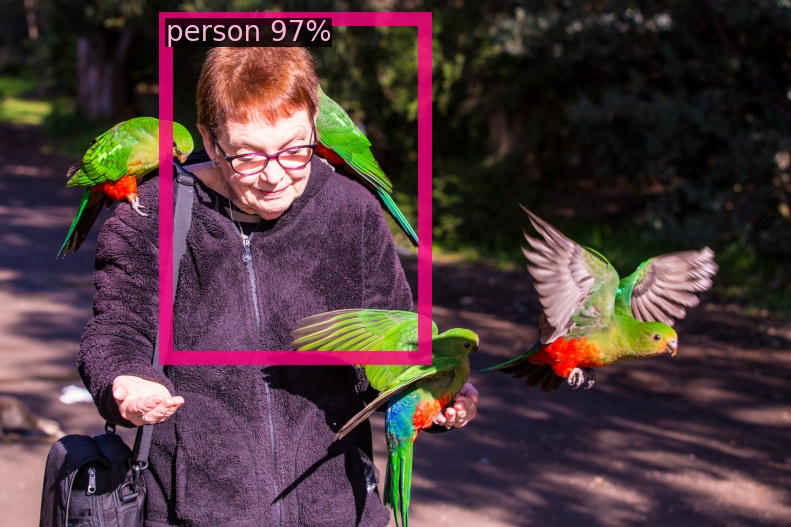}} 
~\subfigure[Incrementally learning $\mathcal{T}_2$]{\includegraphics[width=0.32\linewidth]{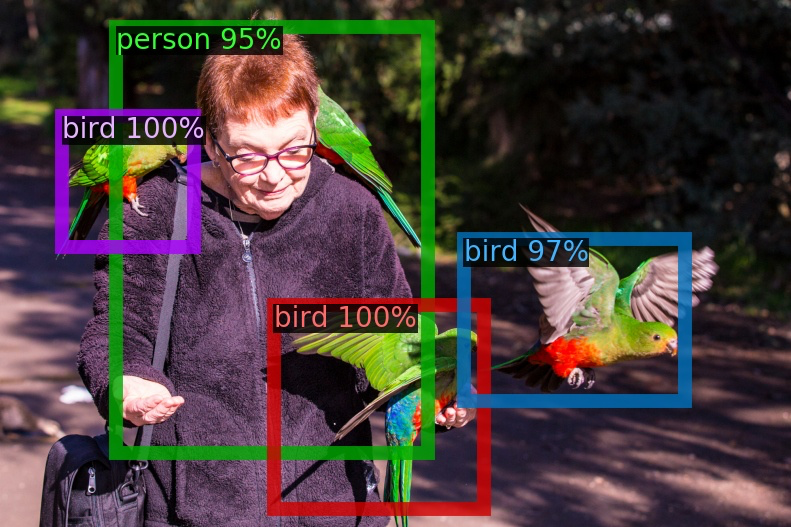}} \vspace{-1em}
\caption{\small 
(a) In Task 1 ($\mathcal{T}_1$), a standard object detector (Faster R-CNN \cite{ren2015faster}) is trained to detect the \texttt{`bird'} class and it can accurately detect \texttt{`bird'} instances on a test image. (b) In Task 2 ($\mathcal{T}_2$), the same model is trained to detect \texttt{`person'} class and it accurately detects a \texttt{`person'} instance. However, the detector forgets the \texttt{`bird'} class ($\mathcal{T}_1$) which was not present during $\mathcal{T}_2$ training. (c) Our meta-learning based incremental Faster R-CNN detector accurately detects most instances of \textit{both} the classes. 
}
\label{fig:intro}\vspace{-12pt}
\end{figure*}


We propose a methodology that views incremental object detection through the lens of meta-learning that can effectively deal with the above challenges. Our meta-learning procedure learns to modify the gradients such that quick adaptation across multiple incremental learning tasks is possible. This is efficiently realised by meta-learning a set of gradient preconditioning matrices, interleaved between layers of a standard object detector (Sec.~\ref{sec:meta_learn_warp_layers}). Further, we formulate a meta-training objective to learn these gradient preconditioning matrices (Sec.~\ref{sec:warp_loss}). In this way, our meta-training procedure captures properties of all the incrementally presented task distributions, effectively alleviating forgetting and preparing the networks for quick adaptation. 
 
\noindent The key contributions of our work  are: 

\begin{itemize}[leftmargin=*,topsep=0pt, noitemsep]
    \item We propose a gradient-based meta-learning approach which learns to reshape gradients such that optimal updates are achieved for both the old and new tasks, for class-incremental object detection problem.  
    \item We propose a novel loss formulation that counters the intransigence enforced due to knowledge distillation, by learning a generalizable set of gradient directions that alleviates forgetting and improves adaptability.
    \item Our extensive evaluations on two benchmark datasets against three competitive baseline methods shows the utility of our meta-learning based methodology.
\end{itemize}

\section{Related Work}
Our proposed methodology lies at the intersection of incremental learning and meta-learning. Hence, we review the literature from both these paradigms here.

\noindent \textbf{Incremental Learning:}
In comparison to the incremental setting for image-classification \cite{kirkpatrick2017overcoming,rebuffi2017icarl,rajasegaran2019random,kj2020meta,rajasegaran2020itaml}, class-incremental object detection and class-incremental semantic segmentation \cite{cermelli2020modeling,michieli2019incremental} has been relatively less explored. Most of the methods proposed so far \cite{shmelkov2017incremental,li2019rilod,hao2019end,chen2019new} use knowledge distillation \cite{hinton2015distilling} to address catastrophic forgetting. These methods mainly vary in the base object detector or parts of the network that are distilled.  
Shmelkov \etal~\cite{shmelkov2017incremental} proposed an incremental version of Fast R-CNN \cite{girshick2015fast}, which uses pre-computed Edge Box object proposal algorithm \cite{zitnick2014edge}, making the setting simpler than what we consider, where the proposal network is also learned. While training for a new task, the classification and regression outputs are distilled from a copy of the model trained on the previous task.  
Li \etal~\cite{li2019rilod} and Chen \etal~\cite{chen2019new} proposed to distill the intermediate features, along with the network outputs. 
Hao \etal~\cite{hao2019end} expands the capacity of the proposal network along with distillation. 
Despite all these efforts, the methods fail to improve the benchmark \cite{shmelkov2017incremental} on the standard evaluation criteria, which calls for thoughts on the effectiveness of the distillation methods and the complexity of the task at hand. 
More recently, Acharya \etal \cite{acharya2020rodeo} proposed to used memory replay to solve a newly formulated online object detection setting, while Joseph \etal~\cite{joseph2021open} explicitly characterised the unknown objects, which was found to be effective in alleviating forgetting.
Peng \etal \cite{peng2020faster} improved the state-of-the-art by using Faster R-CNN with the distillation methodology from \cite{shmelkov2017incremental}. Zhang \etal  \cite{zhang2020class} proposed to distill a consolidated model from a base and incremental detector, using unlabelled auxiliary data.  
In this work, we hypothesise that the restraining effect of distillation may lead to hindrance towards learning new tasks. Therefore to learn a better incremental object detector, the distillation must be carefully modulated with learning for generalizability to new-tasks. To this end, meta-learning offers an attractive solution. 

\noindent \textbf{Meta Learning:}
Meta-learning algorithms can be broadly classified into optimization based methods \cite{finn2017model,nichol2018first,rusu2018meta}, black-box adaptation methods \cite{santoro2016meta,mishra2017simple} and non-parametric methods \cite{snell2017prototypical,sung2018learning}. Adapting these meta-learning methods, which are successful in few-shot image classification setting, to object detection is not straightforward for the reasons enumerated in Sec.~\ref{sec:introduction}. Recently, meta-learning has been applied to address $k$-shot object detection setting. Wang~\etal
\cite{wang2019meta} learn a meta-model that predicts the weights of the RoI Head, that is finally fine-tuned.
Yan~\etal \cite{yan2019meta} and Kang~\etal \cite{kang2019few} learn to re-weight the RoI features and backbone features of Faster R-CNN \cite{ren2015faster} and YOLO \cite{redmon2016you}, respectively. 
Unlike these methodologies, we adopt to use a gradient based meta-learning technique to tackle incremental object detection. 
Inspired by Flennerhag~\etal \cite{Flennerhag2020Meta-Learning}, we propose to meta-learn a gradient preconditioning matrix that encapsulates information across multiple learning tasks. Through extensive experiments, we show that the proposed approach is effective in learning a detector that can be continually adapted to handle new classes.

\section{Methodology} \label{sec:methodology}
The standard object detection frameworks \cite{girshick2014rich,girshick2015fast,redmon2016you,lin2017focal} can be characterised as a function ($\mathcal{F}_{OD}$), that takes an input image and transforms it into a set of bounding boxes enclosing objects, each of which is classified into one of the classes, known a priori. $\mathcal{F}_{OD}$ is trained on large amounts of annotated data corresponding to each class, using variants of stochastic gradient descent.
A \textit{class-incremental object detector} relaxes the constraint that all the class data is available beforehand. As and when new class information is available, the detector should modify itself to be competent on detecting the new classes along with the old classes
by combating itself against catastrophic forgetting \cite{french1999catastrophic,mccloskey1989catastrophic}. 

We formally define the problem in Sec.~\ref{sec:problem_formulation}, introduce how we meta-learn the gradient preconditioning matrix in Sec.~\ref{sec:meta_learn_warp_layers} and finally explain the specifics of our proposed incremental object detector in Sec.~\ref{sec:incr_obj_det}.

\begin{figure*}[h]
\centering
\includegraphics[scale=0.27]{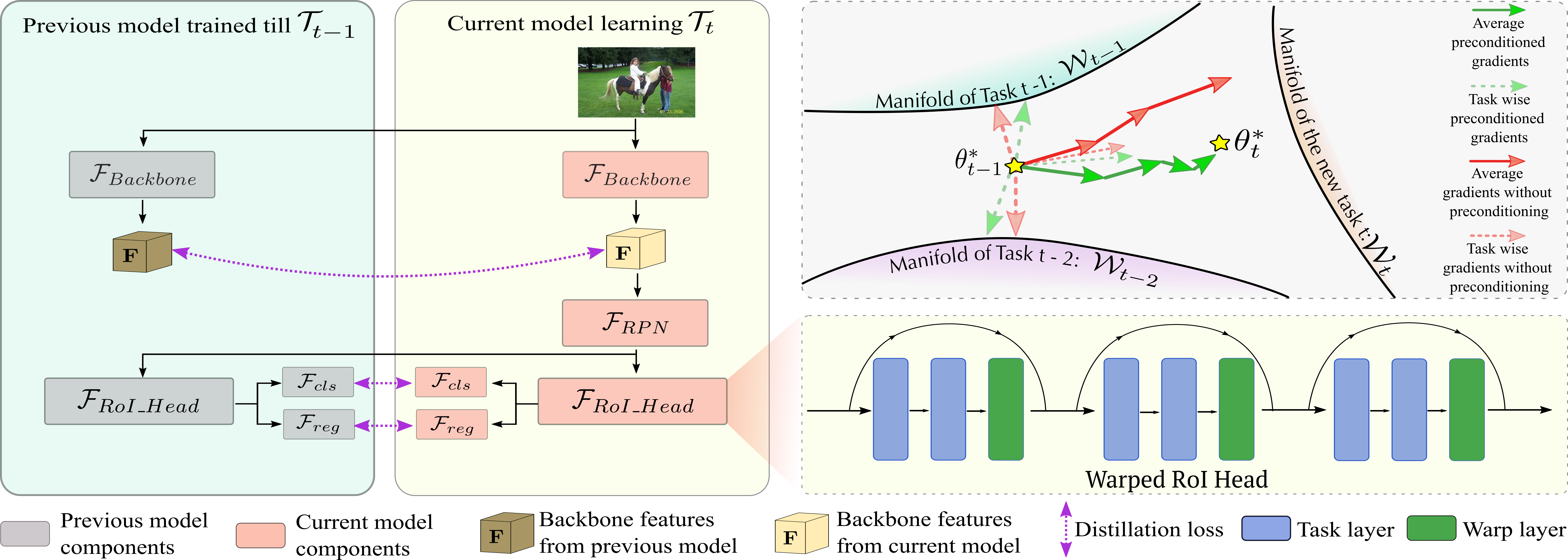}
\vspace{-1.0em}
\caption{\small 
The figure outlines the overall architectural components and an illustration of how gradient preconditioning controls learning (\emph{top right}). While learning the current task $\mathcal{T}_t$, the gradient preconditioning induced by the warp layers (green rectangles in the RoI Head of the detector)
effectively modulates the dotted-red task-wise gradients to the dotted-green gradients, which inherently guides the average gradients (solid green arrows) to a better optima that respects all three task manifolds $\mathcal{W}_{t}, \mathcal{W}_{t-1}$ and $\mathcal{W}_{t-2}$. 
Additionally, backbone features and RoI Head outputs are distilled (purple arrows) from the previous model. \emph{(best viewed in color)}
}
\label{fig:architecture}\vspace{-0.5em}
\end{figure*}
\subsection{Problem Formulation} \label{sec:problem_formulation}
Let $\mathcal{C}$ denotes the set of classes that are incrementally introduced to the object detector. 
A task $\mathcal{T}_t$, is defined as a grouping of these classes, which are exposed to the detector at time $t$: $\mathcal{T}_t \subset \mathcal{C}$, such that $\mathcal{T}_i \cap \mathcal{T}_j = \O $, for any $i, j \leq t$. Let $\mathcal{D}_t$ denote the images containing annotated objects of classes in $\mathcal{T}_t$. Each image can contain multiple objects of different classes, however annotations are available only for those object instances that belong to classes in $\mathcal{T}_t$.

Let $\boldsymbol I \in \mathcal{D}_t$ denotes an input image.  An object detector $\mathcal{F}_{OD}(\boldsymbol I)$, is a composition of functions: $(\mathcal{F}_{RoI\_Head} \circ \mathcal{F}_{RPN}  \circ \mathcal{F}_{Backbone}) (\boldsymbol I)$. Here, without loss of generality, we focus on R-CNN family \cite{girshick2014rich,girshick2015fast,ren2015faster} of two-stage detectors. 
$\mathcal{F}_{Backbone}$ takes $\boldsymbol I$ as input and generates a feature map $\boldsymbol F \in \mathbb{R}^{C\times H \times W}$ where $C, H \text{ and } W$ refer to the number of channels, height and width of the feature map respectively. $\mathcal{F}_{RPN}$ takes these features and proposes areas which can possibly contain an object. Concretely, it outputs a class-agnostic objectness score and the bounding box location for $N$ proposals. Each of these proposals (which is RoI pooled feature from $\boldsymbol F$)
is classified into one of $\mathcal{T}_{i\leq t}$ classes
and its bounding box locations are regressed 
by $F_{RoI\_Head}$. 
Let $\mathcal{F}_{OD}$ be parameterised by  $\bm{\theta}$.
The challenge in class incremental object detection, is to continually adapt $\bm{\theta}$ to learn new tasks $\mathcal{T}_ {t + 1}$, without access to all of \{$\mathcal{D}_{0} \cdots \mathcal{D}_{t}$\}, while maintaining original performance on \{$\mathcal{T}_0 \cdots \mathcal{T}_{t}$\}.

\subsection{Meta-Learning the Gradient Preconditioning} \label{sec:meta_learn_warp_layers}
The standard update rule while training an object detector $\mathcal{F}_{OD}$ parameterised by $\bm{\theta}$ is: $\bm{\theta}^\prime \leftarrow \bm{\theta} - \mu \nabla \mathcal{L}(\bm{\theta})$, where $\mathcal{L}$ is the loss function and $\mu$ is the learning rate. We intend to meta-learn a parameterised preconditioning matrix $P(\bm{\theta}; \bm{\phi})$, which warps the gradient to the steepest direction accounting for the different tasks 
introduced to the object detector till then. The parameter update is thus given by: $\bm{\theta}^\prime \leftarrow \bm{\theta} - \mu P(\bm{\theta}; \bm{\phi}) \nabla \mathcal{L}(\bm{\theta})$, where $\bm{\phi}$ are parameters of $P$. 

A scalable way to achieve such a gradient preconditioning is to embed it directly into the task learner \cite{Flennerhag2020Meta-Learning,lee2017meta,desjardins2015natural}. Following Flennerhag \etal\ \cite{Flennerhag2020Meta-Learning}, we dedicate some of the network layer for preconditioning (called warp layers), which are meta-learned across incremental learning tasks. Hence, the parameter set $\bm{\theta}$ of $\mathcal{F}_{OD}$ is split into task parameters $\bm{\psi}$ and warp parameters $\bm{\phi}$, such that $\bm{\theta} = \bm{\psi} \cup \bm{\phi}$ and $\bm{\psi} \cap \bm{\phi} = \O$. 
During back-propagation, gradient preconditioning is inherently induced on the task layers by the Jacobians of the warp-layers. As the warp layers are non-linear transformations,
it allows modelling rich relationships between gradients. This allows for stronger representational capability than previous works that consider preconditioning via block diagonal matrices \cite{park2019meta,lee2018gradient}.

The warp parameters which precondition the gradients are learned using information from all the tasks seen till then. The carefully defined loss function $\mathcal{L}_{warp}$ (Eqn.~\ref{eqn:warp_loss}) helps achieve this objective. These layers implicitly help to model the joint task distribution of all the tasks introduced to the detector. 
This results in better generalization to new tasks, faster convergence and alleviates catastrophic forgetting.
The illustration in Fig.~\ref{fig:architecture} (top right) explains how gradient preconditioning via warp layers controls learning. Before learning $\mathcal{T}_t$, let $\theta^*_{t-1}$ be the optimal parameters that lie close to the previously learned task manifolds: $\mathcal{W}_{t-1}$ and $\mathcal{W}_{t-2}$. 
While learning the new task $\mathcal{T}_t$, the preconditioning layers effectively warp the task-wise gradients. The transformation of the red dotted arrows to green dotted arrows pictorially shows the change induced by the preconditioning matrix to the task-wise gradients in the first step (the task-wise gradients for the subsequent descent steps are not shown to avoid clutter). Thus, gradient warping helps the average task gradients, shown in green, to converge to a better optima  $\theta_t^*$, which respects all the task manifolds, $\mathcal{W}_{t}, \mathcal{W}_{t-1}$ and $\mathcal{W}_{t-2} $.  


\subsection{Incremental Object Detector} \label{sec:incr_obj_det}
Fig.~\ref{fig:architecture} illustrates the end-to-end architecture of the meta-learned incremental object detector. Faster R-CNN \cite{ren2015faster} is adapted to incorporate additional warp layers which preconditions the gradient flow. The input image is passed through the backbone network ($\mathcal{F}_{Backbone}$) to generate a set of features $\bm F$, which are in turn passed on to the RPN and the RoI pooling layers ($\mathcal{F}_{RPN}$) to generate object features. These are passed to the RoI Head ($\mathcal{F}_{RoI\_Head}$) which contains a set of three residual blocks, each with three convolutional layers. Amongst these convolutional layers, we designate one of them as warp layer (colored  green), which is meta-learned using warp loss, $\mathcal{L}_{warp}$. All the other layers of the network constitute the task layers, which are learned using task loss, $\mathcal{L}_{task}$. $\mathcal{F}_{RoI\_Head}$ terminates in a multi-class classification head ($\mathcal{F}_{cls}$) and a regression head ($\mathcal{F}_{reg}$). While learning a new task, distilling backbone features and final heads from the previous task helps to add additional constraint which guides current learning.
Additional implementation details are discussed in Sec.~\ref{sec:impl_details}.

Below, we explain how $\mathcal{L}_{task}$ and $\mathcal{L}_{warp}$ are formulated, followed by the learning and inference strategies.

\vspace{-0.7em}
\subsubsection{Task Loss ($\mathcal{L}_{task}$):}
A standard object detector is learned by minimising the classification error and the bounding box localization error based on the predictions from the classification and regression heads of $\mathcal{F}_{RoI\_Head}$. Simultaneously, it reduces the discrepancy in objectness score predicted by $\mathcal{F}_{RPN}$ and the corresponding bounding box offsets from the ground-truth. 
Let $\bm{p} = (p_0, \cdots, p_K)$ denote the class probabilities for $K + 1$ classes ($K$ object classes + background class) and let $\bm{l} = (l_x, l_y, l_w, l_h)$ denote the bounding box locations predicted by $\mathcal{F}_{RoI\_Head}$ for each of RoI pooled feature. Let the ground-truth class and bounding box regression targets be  $p^*$ and $\bm{l}^*$. Then, $\mathcal{L}_{RoI\_Head}$ is defined as follows:
\begin{equation}
    \mathcal{L}_{RoI\_Head} = \mathcal{L}_{cls}(\bm{p}, p^*) + \lambda [p^* \geq 1] \mathcal{L}_{loc} (\bm{l}, \bm{l}^*),
    \label{eqn:roi_head}
\end{equation}
where $\mathcal{L}_{cls}(\bm{p}, p^*) = -\log p_{p^*}$ is the log loss for the true class $p^*$ and $\mathcal{L}_{loc}$ is the robust smooth $L_1$ loss function defined in \cite{girshick2015fast}. $p^* = 0$ denotes the background class, and the localisation loss is not calculated for them. 
In a similar manner, the loss for training $\mathcal{F}_{RPN}$ which generates an objectness score $o \in [0,1]$ and the corresponding bounding box predictions $\bm{l}$ is defined as follows:
\begin{equation}
    \mathcal{L}_{RPN} = \mathcal{L}_{cls}(o, o^*) + \lambda o^* \mathcal{L}_{loc}(\bm{l}, \bm{l}^*).
\end{equation}
where $o^*$ is the ground truth which denotes whether the region features contains an object ($= 1$) or not ($= 0$) and $\bm{l}^*$ is the bounding box regression target. The weighting parameter $\lambda$ is set to $1$ for all experiments following \cite{ren2015faster}.

While adapting the current detector $\mathcal{F}_{OD}^{\bm{\theta}_t}$, with parameters $\bm{\theta}_t$ for a new task $\mathcal{T}_t$, we use a frozen copy of the previous model $\mathcal{F}_{OD}^{\bm{\theta}_{t-1}}$, to distill the backbone features and the RoI\_head targets. 
This will ensure that $\mathcal{F}_{OD}^{\bm{\theta}_t}$ does not  deviate too much from $\mathcal{F}_{OD}^{\bm{\theta}_{t-1}}$; which indeed acts as a knowledgeable teacher who has expertise in detecting the previously known classes. 
Each training image is passed through $\mathcal{F}_{Backbone}^{\bm{\theta}_t}$ and $\mathcal{F}_{Backbone}^{\bm{\theta}_{t-1}}$ to obtain $\bm{F}_t$ and $\bm{F}_{t-1}$. Each RoI pooled feature from $\mathcal{F}_{RPN}^{\bm{\theta}_t}$ is passed to $\mathcal{F}_{RoI\_Head}^{\bm{\theta}_t}$ and $\mathcal{F}_{RoI\_Head}^{\bm{\theta}_{t-1}}$ to obtain $\{\bm{p}_t, \bm{l}_t\}$ and $\{\bm{p}_{t-1}, \bm{l}_{t-1}\}$ respectively. 
See purple arrows in Fig.~\ref{fig:architecture} for clarity.
Distillation loss is defined as:
\begin{equation}
    \mathcal{L}_{Distill} = \mathcal{L}_{Reg}(\bm{F}_t, \bm{F}_{t-1}) + \mathcal{L}_{KL}(\bm{p}_t, \bm{p}_{t-1}) + \mathcal{L}_{Reg}(\bm{l}_t, \bm{l}_{t-1}), 
\end{equation}
where $\mathcal{L}_{Reg}$ is the $L_2$ regression loss and $\mathcal{L}_{KL}$ is the KL divergence between the probability distribution generated by  current and previous classification heads, computed only for the previously seen classes. The final task loss is a convex combination of the detection and the distillation losses,
\begin{equation}
    \mathcal{L}_{task} = \alpha \mathcal{L}_{Distill} + (1 - \alpha) (\mathcal{L}_{RPN} + \mathcal{L}_{RoI\_Head})
    \label{eqn:task_loss}
\end{equation}
where $\alpha$ is the weighting factor which controls the importance of each of the term in the loss function. We run a sensitivity analysis on the value of $\alpha$ in Sec.~\ref{sec:ablations}.

\vspace{-0.7em}
\subsubsection{Warp Loss ($\mathcal{L}_{warp}$):} \label{sec:warp_loss}
Motivated by the incremental learning methods for image classification \cite{rebuffi2017icarl,castro2018end}, we maintain an Image Store ($I_{Store}$) with a small number of exemplar images per class. 
We make use of the images from $I_{Store}$ and the current model parameters $\bm{\theta}_t$, to define the warp loss $\mathcal{L}_{warp}$. 
We meta-learn the warp layers in $\mathcal{F}_{RoI\_Head}$ (refer Fig.~\ref{fig:architecture}), which classifies each of the input RoI pooled features into one of the classes and predicts its bounding box locations, using $\mathcal{L}_{warp}$. 
Though $I_{Store}$ guarantees a lower bound on the number of images per class, the actual per class instance statistics would be much uneven. This is because each image can contain multiple instances of different classes. This will heavily bias the warp layer training towards those classes which has more instances. 
To combat this, we propose to use a feature store $F_{Store}$, which stores $N_{feat}$ features per class. This is realised by maintaining a fixed size queue per class and queuing the RoI pooled features into the corresponding class specific queue. This ensures that even if there are multiple instances of many classes in the training data of the detector, the warp layers are updated with equal priority to all the classes. This is a key component that ensures that the gradient preconditioning that is meta-learned would retain equal importance to all the classes. Further, as $I_{Store}$ contains images and annotations from all the classes seen till then, this implicitly embeds information from not only the current class but also the previous classes into the preconditioning layers and therein the whole network, via the warped gradients, effectively reducing forgetting. 

Let $\bm{f}$ denote a single RoI Pooled feature and $p^*$ and $\bm l^*$ denote the corresponding true class label and the bounding box annotation. $\bm f$ is passed through the RoI head $\mathcal{F}_{RoI\_Head}$, to generate class predictions $\bm p$, and box predictions $\bm l$. Then, the warp loss $\mathcal{L}_{warp}$ from the features and labels stored in $F_{Store}$ is computed as,  
\begin{align}
        \mathcal{L}_{warp} 
        =~ & \displaystyle \sum_{(\bm f, p^*, \bm l^*) \sim F_{Store}} \mathcal{L}_{cls}(\bm{p}, p^*) +  [p^* \geq 1] \mathcal{L}_{loc} (\bm{l}, \bm{l}^*), \, \notag \\
        &~ s.t., (\bm p, \bm l) = F_{RoI\_Head}(\bm{f})
    \label{eqn:warp_loss}
\end{align}
Here, $\mathcal{L}_{cls}$ is the log loss and $\mathcal{L}_{loc}$ is a smooth $L_1$ regression loss.
Algorithm \ref{algo:get_warp_loss} illustrates the warp loss computation. Each image in $I_{Store}$ is passed through $\mathcal{F}_{Backbone}$ and $\mathcal{F}_{RPN}$ to generate the RoI pooled features and the associated labels, which is then queued into $F_{Store}$ (Lines 3 - 4). Once all the RoI features are extracted, we accumulate per RoI loss to compute the warp loss $\mathcal{L}_{warp}$ (Lines 6 - 10).

\vspace{-0.8em}
\begin{algorithm}
\caption{Algorithm \textsc{GetWarpLoss}}
\label{algo:get_warp_loss}
\begin{algorithmic}[1]
\Require{Image store: $I_{Store}$; Current model params: $\bm{\theta}_t = \bm{\psi}_t \cup \bm{\phi}_t$. 
}
\State Initialise $F_{Store}$
\Comment{\textit{A queue of length $N_{feat}$ per class}}
\For{$\mathcal{I} \in I_{Store}$}
\State \{($f_{ROI\_pooled}$, labels)\} $\leftarrow \mathcal{F}_{RPN}(\mathcal{F}_{Backbone}(\mathcal{I}))$
\State Enqueue $F_{store}$ with \{($f_{ROI\_pooled}$, labels)\}
\EndFor
\State $\mathcal{L}_{warp} \leftarrow 0$
\For{$f, labels \in F_{Store}$}
\State $p^*, \bm l^* \leftarrow labels$
\State $\bm p, \bm l \leftarrow \mathcal{F}_{RoI\_Head}(f)$
\State $\mathcal{L}_{per\_RoI} = \mathcal{L}_{cls}(\bm{p}, p^*) + [p^* \geq 1] \mathcal{L}_{loc} (\bm{l}, \bm{l}^*)$\Comment{\textit{Eq.~\ref{eqn:warp_loss}}}
\State $\mathcal{L}_{warp} \leftarrow \mathcal{L}_{warp} + \mathcal{L}_{per\_RoI}$
\EndFor
\State \Return $\mathcal{L}_{warp}$
\end{algorithmic}
\end{algorithm}

\subsubsection{Learning and Inference:}
Algorithm \ref{algo:learning-a-task} summarises the end-to-end learning strategy. A mini-batch of datapoints $\mathcal{D}_{tr}$ from the current task is sampled from $\mathcal{D}_t$ in Line 3. 
As introduced in Sec. \ref{sec:warp_loss}, we maintain an Image Store ($I_{Store}$) with a small number of exemplar images per class.
Specifically, $I_{Store}$ contains one fixed size ($N_{img}$) queue per class, which is used to meta-learn the wrap layers. The fixed size queue ensures that only the recently seen $N_{img}$ images per class would be maintained in the store. Images from $\mathcal{D}_{tr}$ are added to $I_{Store}$ in Line 5. For images with multiple class objects in it, we associate it with one of its randomly chosen constituent class. 
The task loss $\mathcal{L}_{task}$ is computed using Eq.~\ref{eqn:task_loss} and the task parameters $\bm{\psi}_t$ are updated in Lines 6 and 7. The warp parameters are updated with warp loss $\mathcal{L}_{warp}$ after every $\gamma$ iterations using $I_{store}$ in Line 11. 
We re-emphasise that while the task layers are updated, the warp layers a kept fixed, which effectively preconditions the task gradients. These preconditioning matrices (layers) are updated using the warp loss by using all the images in $I_{Store}$, which contains images of all the classes introduced to the detector till then, infusing the global information across classes into the task layers. 
A key aspect of detection setting is its innate incremental nature. Even two images from the same task $\mathcal{T}_t$ may not share all same classes. We exploit this behaviour to learn our warp layers even while training on the first task. 

Similar to other meta-learning approaches \cite{finn2017model,nichol2018first,rusu2018meta}, at inference time, the images in $I_{Store}$ are used to fine-tune the task parameters. In this process, meta-learned preconditioning matrix effectively guides the gradients in the steepest descent direction resulting in quick adaptation. In our experiments, we find that with just $10$ examples per task, the model exhibits superior performance. The rest of the inference pipeline follows standard Faster R-CNN \cite{ren2015faster}.
\vspace{-5pt}
\begin{algorithm}
\caption{Learning a current task $\mathcal{T}_t$}
\label{algo:learning-a-task}
\begin{algorithmic}[1]
\Require{Current model params: $\bm{\theta}_t = \bm{\psi}_t \cup \bm{\phi}_t$; Previous model params: $\bm{\theta}_{t-1} = \bm{\psi}_{t-1} \cup \bm{\phi}_{t-1}$; Data for $\mathcal{T}_t$: $\mathcal{D}_t$;  Image store: $I_{Store}$; Warp update interval: $\gamma$; Step len: $\mu, \upsilon$.}
\While{until required iterations}
\State $i \leftarrow 0$
\State $\mathcal{D}_{tr} \leftarrow$ Sample a mini-batch from $\mathcal{D}_t$
\For{ $\mathcal{I} \in \mathcal{D}_{tr}$}
\State Add $\mathcal{I}$ to $I_{Store}$
\State $\mathcal{L}_{task} \leftarrow $ Compute using Eq.~\ref{eqn:task_loss}
\State $\bm{\psi}_t \leftarrow \bm{\psi}_t - \mu \nabla \mathcal{L}_{task}$ \Comment{\textit{Task-parameters update}}
\State $i \leftarrow i + 1$
\If{$i \% \gamma == 0$}
\State $\mathcal{L}_{warp} \leftarrow$ GetWarpLoss($\bm{\theta}_t, I_{store}$)\Comment{\textit{Alg. \ref{algo:get_warp_loss}}}
\State $\bm{\phi}_t \leftarrow \bm{\phi}_t - \upsilon \nabla \mathcal{L}_{warp}$ \Comment{\textit{Meta-parameters update}}
\EndIf
\EndFor
\EndWhile
\State \Return $\bm{\theta}_t, I_{Store}$
\end{algorithmic}
\end{algorithm}







\section{Experiments and Results} \label{sec:expr_res}
We evaluate the proposed approach on a variety of class incremental settings across two prominent detection datasets. We compare against the state-of-the-art methods, consistently outperforming them in all the settings. 
Below, we introduce the datasets, explain the experimental settings, provide implementation details and report our results.

\subsection{Datasets and Evaluation Metrics}
To benchmark our method, we evaluate on PASCAL VOC 2007 \cite{everingham2010pascal} and MS COCO 2014 \cite{lin2014microsoft} datasets following Shmelkov~\etal~\cite{shmelkov2017incremental}.
PASCAL VOC 2007 contains $9{,}963$ images containing $24{,}640$ annotated instances of 20 categories. Following the standard setting \cite{everingham2010pascal}, $50 \%$ of data is split into train/val splits and the rest for testing on PASCAL VOC. MS COCO 2014 contains objects from 80 different categories with $83{,}000$ images in its training set and $41{,}000$ images in the validation set. Since the MS COCO test set is not available, we use the validation set for evaluation. 

The mean average precision at $0.5$ IoU threshold (mAP@50) is used as the primary evaluation metric for both datasets. For MS COCO, we additionally report average precision and recall across scales, the number of detections and IoU thresholds, in line with its standard protocol. 
\begin{table*}[h]
\centering
\resizebox{\textwidth}{!}{%
\begin{tabular}{@{}l|l|aaaaaaaaaaaaaaakbgdef|l@{}}
\toprule
Class Split                                                                 &          & aero          & cycle         & bird          & boat          & bottle        & bus           & car           & cat           & chair         & cow           & table         & dog           & horse         & bike          & person        & \underline{mAP-old}       & plant         & sheep         & sofa          & train         & tv            & \underline{mAP}           \\ \midrule
1-20                                                                        &          & 79.4          & 83.3          & 73.2          & 59.4          & 62.6          & 81.7          & 86.6          & 83.0          & 56.4          & 81.6          & 71.9          & 83.0          & 85.4          & 81.5          & 82.7        & 76.8  & 49.4          & 74.4          & 75.1          & 79.6          & 73.6          & 75.2          \\
1-15                                                                        &          & 78.1          & 82.6          & 74.2          & 61.8          & 63.9          & 80.4          & 87.0          & 81.5          & 57.7          & 80.4          & 73.1          & 80.8          & 85.8          & 81.6          & 83.9        & 76.9  & -             & -             & -             & -             & -             & 76.9          \\ \midrule
\multirow{2}{*}{(1-15)+16}                                                  & \cite{shmelkov2017incremental} & 70.5          & 78.3          & 69.6 & 60.4 & 52.4          & 76.8  & 79.4          & 79.2          & 47.1          & 70.2          & 56.7          &  77.0 & 80.3          &  78.1 & 70.0      & 69.7    & 26.3          & -             & -             & -             & -             & 67.0          \\
                                                                            & Ours     & 78.8 & 79.0 & 65.9          & 51.8          & 57.3 & 76.1          & 84.2 & 80.3 & 47.3 & 77.0 & 60.4 & 76.0          & 81.8 & 76.9          & 80.6 & \textbf{71.6} & 33.3 & -             & -             & -             & -             & \textbf{69.2} \\ \midrule
\multirow{2}{*}{\begin{tabular}[c]{@{}l@{}}(1-15)+\\ 16+17\end{tabular}}    &  \cite{shmelkov2017incremental} & 70.3          & 78.9          & {67.7} & {59.2} & 47.0          & 76.3          & 79.3          & 77.7          & {48.0} & 58.8          & 60.2          & 67.4          & 71.6          & {78.6} & 70.2        & 67.4  & 27.9          & 46.8          & -             & -             & -             & 63.9          \\
                                                                            & Ours     & {80.1} & {79.9} & 66.2          & 53.4          & {58.4} & {79.6} & {84.9} & {79.0} & 47.3          & {72.9} & {60.7} & {73.7} & {81.2} & 78.3          & {80.4} & \textbf{71.7} & {35.5} & {56.3} & -             & -             & -             & \textbf{68.7} \\ \midrule
\multirow{2}{*}{\begin{tabular}[c]{@{}l@{}}(1-15)+\\ 16+..+18\end{tabular}} &  \cite{shmelkov2017incremental} & 69.8          & 78.2          & {67.0} & 50.4          & 46.9          & 76.5          & 78.6          & {78.0} & {46.4} & 58.6          & 58.6          & 67.5          & 71.8          & {78.5} & 69.9         & 66.4 & 26.1          & {56.2}          & 45.3          & -             & -             & 62.5          \\
                                                                            & Ours     & {80.1} & {79.4} & 65.0          & {53.2} & {58.0} & {78.4} & {85.0} & 77.5          & 46.3          & {72.9} & {59.2} & {74.1} & {81.4} & 75.7          & {79.9} & \textbf{71.1} & {35.4} & 55.8          & {50.0} & -             & -             & \textbf{67.1} \\ \midrule
\multirow{2}{*}{\begin{tabular}[c]{@{}l@{}}(1-15)+\\ 16+..+19\end{tabular}} &  \cite{shmelkov2017incremental} & 70.4          & {78.8} & {67.3} & 49.8          & 46.4          & {75.6} & 78.4          & 78.0            & 46.0          & 59.5          & 59.2          & 67.2          & 71.8          & 71.3          & 69.8         & 66 & 25.9          & 56.1          & 48.2          & 65.0          & -             & 62.4          \\
                                                                            & Ours     & {77.7} & 78.5          & 67.1          & {51.9} & {56.9} & 74.9          & {84.1} & {79.1} & {46.4} & {71.8} & {59.4} & {72.5} & {81.6} & {75.6} & {79.5} & \textbf{70.5} & {33.5} & {58.2} & {49.4} & {65.5} & -             & \textbf{66.5} \\ \midrule
\multirow{2}{*}{\begin{tabular}[c]{@{}l@{}}(1-15)+\\ 16+..+20\end{tabular}} &  \cite{shmelkov2017incremental} & 70.0          & {78.1} & 61.0          & 50.9          & 46.3          & {76.0} & 78.8          & {77.2} & {46.1} & 66.6          & {58.9} & 67.7          & 71.6          & 71.4          & 69.6       & 66  & 25.6          & 57.1          & 46.5          & {70.7} & {58.2} & 62.4          \\
                                                                            & Ours     & {77.5} & 77.1          & {66.8} & {53.6} & {55.0} & 73.7          & {83.6} & 76.7          & 45.2          & {74.0} & 57.7          & {72.4} & {81.3} & {77.0} & {79.1} & \textbf{70.0} &  {34.9} & {58.1} & {49.6} & 67.5          & 53.7          & \textbf{65.7} \\ \bottomrule
\end{tabular}%
}\vspace{0.3em}
\caption{\small Per-class AP and overall mAP values when five classes from PASCAL VOC dataset are added one-by-one to a model initially trained on 15 categories on the left. `mAP-old' column refers to the mAP of all the base classes, while `mAP' column refers to the mAP of all the classes seen till then. Our approach achieves consistent improvement in detection performance, compared to~\cite{shmelkov2017incremental}.
}
\label{tab:one_plus_one}
\vspace{-22pt}
\end{table*}

\vspace{-6pt}
\subsection{Experimental Settings} \label{sec:experimental_setting}
Following \cite{shmelkov2017incremental}, we simulate incremental versions of both PASCAL VOC and MS COCO datasets. 
As introduced in Sec.~\ref{sec:problem_formulation}, a group of classes constitute a task $\mathcal{T}_t$, which is presented to the learner at time $t$. Let $C$ denotes the set of classes that are part of $\mathcal{T}_t$. The data  $\mathcal{D}_t$ for task $\mathcal{T}_t$ is created by selecting those images which have any of the classes in $C$. Instances of those classes that are not part of $C$, but still co-occur in a selected image would be left unlabelled. 

For PASCAL VOC, we order the classes alphabetically and create multiple tasks by grouping them. We consider four different settings, in the decreasing order of difficulty: (a) the first task $\mathcal{T}_1$ containing initial $15$ classes and the next five successive tasks ($\mathcal{T}_2 \cdots \mathcal{T}_6$) containing a new class each. (b) $\mathcal{T}_1$ containing first $15$ classes and $\mathcal{T}_2$ containing the rest of $5$ classes. (c) $\mathcal{T}_1$ containing first $10$ classes and $\mathcal{T}_2$ containing the other $10$ classes. (d) Grouping all the initial $19$ classes in $\mathcal{T}_1$ and the last class into $\mathcal{T}_2$. For MS COCO, we use the first $40$ classes as task $\mathcal{T}_1$ and the rest as $\mathcal{T}_2$.

\vspace{-6pt}
\subsection{Implementation Details} \label{sec:impl_details}
We build our incremental object detector based on Faster R-CNN \cite{ren2015faster}. 
Continually learning a Faster R-CNN is more challenging than the case of Fast R-CNN (as deployed in Shmelkov \etal \cite{shmelkov2017incremental}), since it makes use of  pre-computed Edge Box proposals \cite{zitnick2014edge}, while we also learn the RPN.
To maintain fairness in  comparison with \cite{shmelkov2017incremental}, we use ResNet-50 \cite{he2016deep} with frozen batch normalization layers as the backbone.

The classification head of Faster R-CNN handles only the classes seen so far.  
Following other class incremental works \cite{lopez2017gradient,rajasegaran2019random,chaudhry2019efficient}, this is done by setting logits of the unseen class to a very high negative value ($-10^{10}$). This makes the contribution of the unseen classes in the softmax function negligible ($e^{-10^{10}} \rightarrow 0$), while computing the class probabilities (referred to as $p$ in Eq.~\ref{eqn:roi_head}). 
While updating the task parameters $\psi$, the warp parameters $\phi$ are kept fixed and vice-versa. This is achieved by selectively zeroing out the gradients during the backward pass of the corresponding loss functions; $\mathcal{L}_{task}$ and $\mathcal{L}_{warp}$. Other than these two modifications, the architectural components and the training methodology is the same as standard Faster R-CNN.

We use stochastic gradient descent (SGD) with momentum $0.9$. The initial learning rate is set to $0.02$ and subsequently reduced to $0.0002$, with a warm-up period of $100$ iterations. Each task is trained for $18{,}000$ and $90{,}000$ iterations for PASCAL VOC and MS COCO respectively. The training is carried out on a single machine with $8$ GPUs, each of them processing two images at a time. Hence, the effective batch size is $16$. During evaluation, $100$ detections per image are considered with an NMS threshold of 0.4.
$N_{feat}$ and $N_{img}$, which control the queue size of the feature store ($F_{Store}$) and image store ($I_{Store}$) respectively, are set to $10$. 
Note that both $F_{Store}$ and $I_{Store}$ are class specific queues of fixed length. Hence they do not grow in size as the oldest item will be dequeued as a new item is enqueued to an already full queue. 
The warp update interval $\gamma$ and $\alpha$ in Eq. \ref{eqn:task_loss}, is set to $20$ and $0.2$ respectively.
Our implementation is based on Detectron2 \cite{wu2019detectron2} library. We would be releasing the code and trained models for further clarity and reproducibility. 

\subsection{Results} \label{sec:results}
In the following, we organise the results for each of the experimental settings outlined in Sec.~\ref{sec:experimental_setting}. The classes introduced in each task are color coded for clarity. 

\noindent\textbf{Adding Classes Sequentially:}
In the first experiment, we consider incrementally adding one new class at a time to the object detector that is trained to detect all the previously seen classes.
We simulate this scenario by training the detector on images from the first $15$ classes of PASCAL VOC and then adding $16^{th}-20^{th}$ classes one by one.

Table \ref{tab:one_plus_one} shows the class-wise average precision (AP) at IoU threshold $0.5$ and the corresponding mean average precision (mAP). The first row is the upper-bound where the detector is trained on data from all $20$ classes. The AP values when $\mathcal{F}_{OD}$ is trained on first $15$ class examples and evaluated on test data from the same $15$ classes is reported in the second row. The following five pair of rows showcase the result when a new class is added. The notation $(1-15)+16+..+20$ is a shorthand for this setting. Our  meta-learned model performs favourably well against the previous best method \cite{shmelkov2017incremental} on all the sequential tasks. 

\begin{table*}
\parbox{1\linewidth}{
\centering
\resizebox{\textwidth}{!}{%
\begin{tabular}{@{}l|aaaaaaaaaabbbbbbbbbb|l@{}}
\toprule
Train with & aero        & cycle         & bird          & boat          & bottle        & bus           & car           & cat         & chair         & cow           & table         & dog           & horse       & bike          & person        & plant         & sheep       & sofa          & train         & tv            & mAP           \\ \midrule
All 20     & 79.4        & 83.3          & 73.2          & 59.4          & 62.6          & 81.7          & 86.6          & 83.0          & 56.4          & 81.6          & 71.9          & 83.0            & 85.4        & 81.5          & 82.7          & 49.4          & 74.4        & 75.1          & 79.6          & 73.6          & 75.2          \\
First 10   & 78.6        & 78.6          & 72.0           & 54.5          & 63.9          & 81.5          & 87.0            & 78.2        & 55.3          & 84.4          & -             & -             & -           & -             & -             & -             & -           & -             & -             & -             & 73.4          \\
Std Training    & 35.7 & 9.1   & 16.6 & 7.3  & 9.1    & 18.2 & 9.1  & 26.4 & 9.1   & 6.1  & 57.6  & 57.1 & 72.6  & 67.5 & 73.9   & 33.5  & 53.4  & 61.1 & 66.5  & 57.0   & 37.3 \\ \midrule
Shmelkov \etal \cite{shmelkov2017incremental}  & 69.9        & 70.4          & {69.4} & 54.3          & 48.0            & 68.7          & 78.9          & 68.4        & {45.5} & 58.1          & 59.7          & {72.7} & 73.5        & {73.2} & 66.3          & 29.5          & 63.4        & {61.6} & {69.3} & {62.2} & 63.1          \\
Faster ILOD \cite{peng2020faster} & 72.8 & 75.7 & 71.2 & 60.5 & 61.7 & 70.4 & 83.3 & 76.6 & 53.1 & 72.3 & 36.7 & 70.9 & 66.8 & 67.6 & 66.1 & 24.7 & 63.1 & 48.1 & 57.1 & 43.6 & 62.2 \\
ORE~\cite{joseph2021open} & 63.5 & 70.9 & 58.9 & 42.9 & 34.1 & 76.2 & 80.7 & 76.3 & 34.1 & 66.1 & 56.1 & 70.4 & 80.2 & 72.3 & 81.8 & 42.7 & 71.6 & 68.1 & 77.0 & 67.7 & 64.6 \\ \midrule
Ours       & {76.0} & {74.6} & 67.5          & {55.9} & {57.6} & {75.1} & {85.4} & {77.0} & 43.7          & {70.8} & {60.1} & 66.4          & {76.0} & 72.6          & {74.6} & {39.7} & {64.0} & 60.2          & 68.5          & 60.5          & {\textbf{66.3}} \\
 \bottomrule
\end{tabular}%
}\vspace{0.3em}
\caption{\small Per-class AP and overall mAP on PASCAL VOC when \colorbox{ task2}{10 new classes} are added to a detector trained on the \colorbox{ task1}{first 10 classes.}
}
\label{tab:10_plus_10}
\vspace{-0.5em}
}
\hfill
\parbox{1\linewidth}{
\centering

\resizebox{\textwidth}{!}{%
\begin{tabular}{@{}l|aaaaaaaaaaaaaaabbbbb|l@{}}
\toprule
Train with & aero          & cycle         & bird          & boat          & bottle        & bus           & car           & cat           & chair         & cow         & table         & dog           & horse         & bike          & person        & plant         & sheep         & sofa          & train       & tv            & mAP           \\ \midrule
All 20     & 79.4          & 83.3          & 73.2          & 59.4          & 62.6          & 81.7          & 86.6          & 83.0            & 56.4          & 81.6        & 71.9          & 83.0            & 85.4          & 81.5          & 82.7          & 49.4          & 74.4          & 75.1          & 79.6        & 73.6          & 75.2          \\
First 15   & 78.1          & 82.6          & 74.2          & 61.8          & 63.9          & 80.4          & 87.0            & 81.5          & 57.7          & 80.4        & 73.1          & 80.8          & 85.8          & 81.6          & 83.9          & -             & -             & -             & -           & -             & 53.2          \\
Std Training     & 12.7          & 0.6           & 9.1           & 9.1           & 3.0             & 0.0             & 8.5           & 9.1           & 0.0             & 3.0           & 9.1           & 0.0             & 3.3           & 2.3           & 9.1           & 37.6          & 51.2          & 57.8          & 51.5        & 59.8          & 16.8          \\ \midrule
Shmelkov \etal \cite{shmelkov2017incremental}  & 70.5          & 79.2          & {68.8} & {59.1} & 53.2          & 75.4          & 79.4          & 78.8          & 46.6          & 59.4        & 59.0            & {75.8} & 71.8          & {78.6} & 69.6          & 33.7          & {61.5} & {63.1} & 71.7        & {62.2} & 65.9          \\ 
Faster ILOD \cite{peng2020faster} & 66.5 & 78.1 & 71.8 & 54.6 & 61.4 & 68.4 & 82.6 & 82.7 & 52.1 & 74.3 & 63.1 & 78.6 & 80.5 & 78.4 & 80.4 & 36.7 & 61.7 & 59.3 & 67.9 & 59.1 & 67.9 \\
ORE \cite{joseph2021open} & 75.4 & 81.0 & 67.1 & 51.9 & 55.7 & 77.2 & 85.6 & 81.7 & 46.1 & 76.2 & 55.4 & 76.7 & 86.2 & 78.5 & 82.1 & 32.8 & 63.6 & 54.7 & 77.7 & 64.6 & \textbf{68.5} \\
\midrule
Ours       & {78.4} & {79.7} & 66.9          & 54.8          & {56.2} & {77.7} & {84.6} & {79.1} & {47.7} & {75.0} & {61.8} & 74.7          & {81.6} & 77.5          & {80.2} & {37.8} & 58.0            & 54.6          & {73.0} & 56.1          & {67.8} \\ 
\bottomrule
\end{tabular}%
}\vspace{0.3em}
\caption{\small Per-class AP and overall mAP when \colorbox{ task2}{last 5 classes} from PASCAL VOC are added to a detector trained on the \colorbox{ task1}{initial 15 classes.}}
\label{tab:15_plus_5}
\vspace{-0.5em}
}
\hfill
\parbox{1\linewidth}{
\centering

\resizebox{\textwidth}{!}{%
\begin{tabular}{@{}l|aaaaaaaaaaaaaaaaaaab|l@{}}
\toprule
Train with & aero          & cycle         & bird          & boat          & bottle      & bus           & car           & cat           & chair         & cow         & table         & dog           & horse         & bike          & person        & plant         & sheep         & sofa          & train         & tv            & mAP           \\ \midrule
All 20     & 79.4          & 83.3          & 73.2          & 59.4          & 62.6        & 81.7          & 86.6          & 83.0            & 56.4          & 81.6        & 71.9          & 83.0            & 85.4          & 81.5          & 82.7          & 49.4          & 74.4          & 75.1          & 79.6          & 73.6          & 75.2          \\
First 19   & 76.3          & 77.3          & 68.4          & 55.4          & 59.7        & 81.4          & 85.3          & 80.3          & 47.8          & 78.1        & 65.7          & 77.5          & 83.5          & 76.2          & 77.2          & 46.6          & 71.4          & 65.8          & 76.5          & -             & 67.5          \\
Std Training    & 16.6          & 9.1           & 9.1           & 9.1           & 9.1         & 8.3           & 35.3          & 9.1           & 0.0             & 22.3        & 9.1           & 9.1           & 9.1           & 13.7          & 9.1           & 9.1           & 23.1          & 9.1           & 15.4          & 50.7          & 14.3          \\ \midrule
Shmelkov \etal \cite{shmelkov2017incremental}   & 69.4          & {79.3} & {69.5} & {57.4} & 45.4        & 78.4          & 79.1          & {80.5} & 45.7          & 76.3        & {64.8} & 77.2          & 80.8          & 77.5          & 70.1          & 42.3          & 67.5          & 64.4          & {76.7} & {62.7} & 68.3          \\
Faster ILOD \cite{peng2020faster} & 64.2 & 74.7 & 73.2 & 55.5 & 53.7 & 70.8 & 82.9 & 82.6 & 51.6 & 79.7 & 58.7 & 78.8 & 81.8 & 75.3 & 77.4 & 43.1 & 73.8 & 61.7 & 69.8 & 61.1 & 68.6 \\
ORE \cite{joseph2021open} & 67.3 & 76.8 & 60.0 & 48.4 & 58.8 & 81.1 & 86.5 & 75.8 & 41.5 & 79.6 & 54.6 & 72.8 & 85.9 & 81.7 & 82.4 & 44.8 & 75.8 & 68.2 & 75.7 & 60.1 & 68.9 \\ \midrule
Ours       & {78.2} & 77.5          & 69.4          & 55.0            & {56.0} & {78.4} & {84.2} & 79.2          & {46.6} & {79.0} & 63.2          & {78.5} & {82.7} & {79.1} & {79.9} & {44.1} & {73.2} & {66.3} & 76.4          & 57.6          & {\textbf{70.2}} \\
\bottomrule
\end{tabular}%
}\vspace{0.3em}
\caption{\small Per-class AP and overall mAP when \colorbox{task2}{tvmonitor class} from PASCAL VOC is added to the detector, trained on all \colorbox{task1}{other classes.} }
\label{tab:19_plus_1}
\vspace{-0.5em}
}
\end{table*}

\begin{figure*}[h]
\centering
\vspace{-1em}
\includegraphics[scale=0.12]{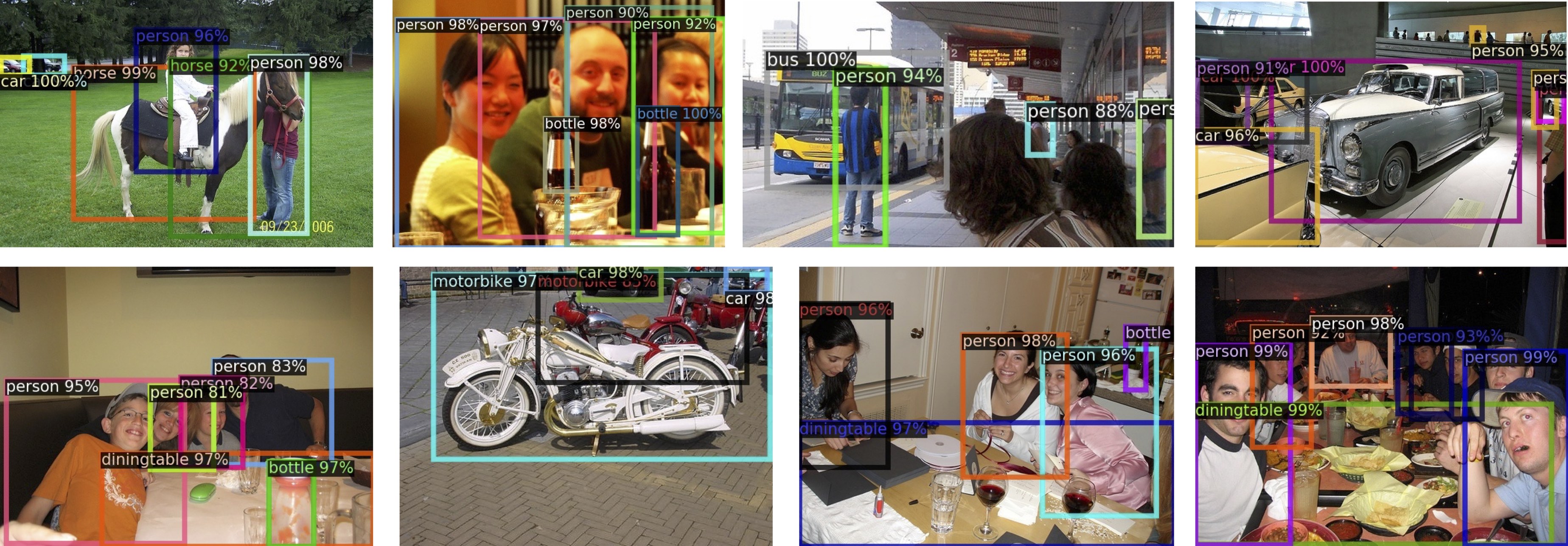}
\vspace{-0.7em}
\caption{\small 
Qualitative results of our incremental object detector trained in a 10+10 setting where $\mathcal{T}_1 = \{$aeroplane, bicycle, bird, boat, bottle, bus, car, cat, chair, cow$\}$ and $\mathcal{T}_2 = \{$diningtable, dog, horse, motorbike, person, pottedplant, sheep, sofa, train, tvmonitor$\}$.
}
\label{fig:qualitative_results}
\vspace{-2pt}
\end{figure*}

\noindent\textbf{Adding Groups of Classes Together:}
Next, we test our method in a dual task scenario, where $\mathcal{T}_1$ contains one set of classes and $\mathcal{T}_2$ contains the remaining classes. We consider $10 + 10$, $15 + 5$ and $19 + 1$ settings for PASCAL VOC.
Tables \ref{tab:10_plus_10}, \ref{tab:15_plus_5} and  \ref{tab:19_plus_1} show the corresponding results. The second row in each of the tables shows the upper-bound when all class data is available for training. The third row reports the AP values when we train and evaluate $\mathcal{F}_{OD}$ on $\mathcal{T}_1$. When the second task $\mathcal{T}_2$ is added with standard training, we see that performance on classes of the first task drops significantly (fourth row). This evaluation is carried out on test examples from all 20 classes.
In the subsequent rows, we report the accuracies of our proposed methodology when compared against Shmelkov \etal\cite{shmelkov2017incremental}, Faster ILOD \cite{peng2020faster} and ORE \cite{joseph2021open}
. We see that our approach comfortably outperforms \cite{shmelkov2017incremental} and Faster ILOD \cite{peng2020faster} in terms of mAP in all the settings. 
ORE \cite{joseph2021open} has a slightly better performance in the 15+5 setting. 
ORE's capability to model unknown objects explicitly is orthogonal to our work and can be incorporated into ours. We will explore this direction in a future work. 
Class-wise AP values are also reported, showing our improvements on majority of the classes.
We compare with DMC \cite{zhang2020class} in the Supplementary.
Qualitative results are showcased in Fig.~\ref{fig:qualitative_results}.

Table \ref{tab:coco_40_plus_40} reports the results on MS COCO in a $40 + 40$ setting (results are reported on entire val-set). For the sake of comparison with \cite{shmelkov2017incremental,peng2020faster}, we also report results on mini-val, which contains the first 5000 images from the validation split. Following the standard COCO evaluation, we report average precision across multiple IoUs ($AP$-(.50:.05:.95), $AP^{50}$-.50, $AP^{75}$-.75) and scales ($AP^{S}$-small, $AP^{M}$-medium and $AP^{L}$-large) and average recall while using 1, 10 and 100 detections per image ($AR^{1}, AR^{10}, AR^{100}$) and different scales ($AR^{S}$-small, $AR^{M}$-medium and $AR^{L}$-large).

\section{Discussions and Analysis} \label{sec:ablations}
Here, we report ablation results and the justification for our design choices.  
All the experiments are conducted in the $15 + 5$ setting, where $5$ new classes are added to the detector trained on the first $15$ classes of PASCAL VOC \cite{everingham2010pascal}.

\noindent\textbf{Ablation Experiment:}
We design a set of experiments to clearly understand the contribution of each of the constituent components of the proposed methodology: distillation, gradient preconditioning and fine-tuning. 
Table \ref{tab:interplay} shows the results of this ablation study where each of these components are selectively switched on (\checkmark) and off ($\times$) in a $15 
+ 5$ setting. 
We note that using only the distillation loss to avoid forgetting during the training for a new task helps to achieve $58.5 \%$ mAP. Just using gradient preconditioning during incremental learning results in a lower mAP of $47.6 \%$, but fine-tuning it with the same amount of data results in $13.5 \%$ mAP improvement when compared to fine-tuning a distilled model, which results in only $5.1 \%$ mAP improvement. This clearly brings out the effectiveness of meta-learning the gradient preconditioning layers, for quick adaptability. Finally, although using both distillation and gradient preconditioning helps to achieve an mAP of $54.3 \%$, the best performance is achieved when fine-tuning is also applied resulting in an mAP of $67.8 \%$. 

\begin{table}[t]\setlength{\tabcolsep}{1pt}
\centering
\resizebox{1\linewidth}{!}{%
\begin{tabular}{@{}p{0.04\textwidth}|l|llllllllllll@{}}
\toprule
\multicolumn{2}{!}{}   & $AP$ & $AP^{50}$ & $AP^{75}$ & $AP^{S}$ & $AP^{M}$ & $AP^{L}$ & $AR^{1}$ & $AR^{10}$ & $AR^{100}$ & $AR^{S}$ & $AR^{M}$ & $AR^{L}$ \\ \midrule
\multirow{4}{1cm}{mini-val} & All 80 & 31.5 & 50.9 & 33.5 & 15.4 & 35.6 & 43.5 & 28.7 & 44.3 & 45.9 & 26.1 & 51.2 & 61.8 \\ \cline{2-14}
 & \cite{shmelkov2017incremental} & 21.3 & 37.4 & - & - & - & - & - & - & - & - & - & - \Tstrut \\ 
 & \cite{peng2020faster} & 20.6 & 40.1 & - & - & - & - & - & - & - & - & - & - \Tstrut \\
 & Ours & \textbf{23.8} & \textbf{40.5} & 24.4 & 12 & 26.6 & 32 & 24.2 & 38.4 & 39.7 & 20.7 & 44.6 & 52.4 \\ \midrule
\multirow{2}{0.8cm}{full-val} & All 80 & 31.2 & 51 & 33.1 & 14.8 & 34.6 & 41.5 & 28.9 & 44.8 & 46.4 & 25 & 51.5 & 61.7 \\ \cline{2-14}
 & Ours & 23.7 & 40.4 & 24.5 & 11.8 & 26.2 & 30 & 24.3 & 38.6 & 40 & 20.5 & 44.3 & 52.7 \Tstrut \\ \bottomrule
\end{tabular}%
}\vspace{0.3em}
\caption{\small AP and AR values when 40 new classes are added to a model trained with the first 40 classes on MS COCO dataset.}
\label{tab:coco_40_plus_40}
\vspace{-5pt}
\end{table}

\noindent\textbf{Choice of Preconditioning Layers:}
We characterise the forgetting in a two stage object detector and identify that the backbone, the RPN ($\mathcal{F}_{RPN}$) and the RoI Head ($\mathcal{F}_{RoI\_Head}$) have different forgetting rates. This can be attributed to the category-agnostic nature of $\mathcal{F}_{RPN}$. It learns to predict regions that can possibly contain an object (expressed by the objectness score), making it independent of the object class. 
We hypothesise that keeping the RPN fixed or training it along for the new classes would not lead to significant performance change. Our experimental results corroborate with this, where we see a similar overall mAP of $58.5 \% $ vs. $58.1 \%$ for a distilled detector with and without trained RPN. 

We choose to add the preconditioning layers in $\mathcal{F}_{RoI\_Head}$ because: (a) Its category-specific nature as opposed to category-agnostic nature of $\mathcal{F}_{RPN}$ makes it the most forgetful component in the object detection pipeline. (b) Since each RoI-pooled feature (which is an input to $\mathcal{F}_{RoI\_Head}$) corresponds to only a single object, this helps to meta-learn the preconditioning layers in a class balanced manner using the $F_{Store}$. (c) The earlier the preconditioning is applied to the gradients in the back-propagation path, the larger portion of the network is adapted effectively. 
We choose one layer from each of the residual blocks in $\mathcal{F}_{RoI\_Head}$ as preconditioning layers (refer Fig.~\ref{fig:architecture}). Empirically, we find that adding the preconditioning layers to the last residual block yields the best result ($58.5 \%$ mAP), when compared to adding it to the other two layers ($55.6 \%$ and $54.1 \%$ mAP respectively).

\begin{table}[t]
\setlength{\tabcolsep}{4pt}
\parbox{.50\linewidth}{
\centering
\resizebox{0.2\textwidth}{!}{
\begin{tabular}{@{}lll|ccc@{}}
\toprule
D          & G          & F          & $\mathcal{T}_1$ & $\mathcal{T}_2$ & All \\ \midrule
\checkmark & $\times$   & $\times$   & 64.5            & 40.8            & 58.5                                     \\
\checkmark & $\times$   & \checkmark & 71.0              & 41.6            & 63.6                                     \\
$\times$   & \checkmark & $\times$   & 45.6            & 53.6            & 47.6                                     \\
$\times$   & \checkmark & \checkmark & 69.7            & 55.2            & 66.1                                     \\
\checkmark & \checkmark & $\times$   & 58.2            & 42.7            & 54.3                                     \\
\checkmark & \checkmark & \checkmark & 71.7            & 55.9            & 67.8                                     \\ \bottomrule
\end{tabular}}\vspace{0.2em}
\caption{An ablation study to understand the contribution of Distillation (D), Gradient preconditioning (G) and  Fine-tuning (F).}
\label{tab:interplay}
}
\hfill
\parbox{.48\linewidth}{
\centering \setlength{\tabcolsep}{5pt}
\resizebox{0.17\textwidth}{!}{
\begin{tabular}{@{}l|ccc@{}}
\toprule
$\alpha$ & $\mathcal{T}_1$ & $\mathcal{T}_2$ & All   \\ \midrule
0.1          & 64.5   & 40.8   & 58.54 \\
0.2          & 66.5   & 35.2   & 58.67 \\
0.4          & 68.5   & 27.0     & 58.08 \\
0.6          & 69.1   & 16.8   & 56.03 \\
0.8          & 70.0     & 3.0      & 53.23 \\ \bottomrule
\end{tabular}}\vspace{0.2em}
\caption{Sensitivity analysis on the hyper-parameter $\alpha$, in Eq.~\ref{eqn:task_loss}.  $\alpha$ controls the importance of distillation and detection losses.}
\label{tab:sensitivity_analysis}
}
\vspace{-5pt}
\end{table}

\noindent\textbf{Sensitivity Analysis:}
We run a sensitivity analysis on the weighting factor $\alpha$ which weighs the importance of the distillation loss and the object detection loss in Eq.~\ref{eqn:task_loss}. The results are reported in Table \ref{tab:sensitivity_analysis}, where we clearly see that increasing the importance for distillation, reduces the ability to learn new tasks. 
Only distillation technique is used to reduce forgetting in these results.

\noindent\textbf{Choice of Feature Distillation Methodology:}
We explore two other distillation methods than using simple $L_2$ loss. 
Neuron Selectivity Transfer (NST) \cite{huang2017like} uses Maximum Mean Discrepancy (MMD) to measure and match the features of the source and the target models, while Activation-based attention Transfer (AT) \cite{zagoruyko2016paying} forces the target network to mimic the attention maps of the source network.
While both the methods seem appealing, their performance is significantly lower than $L_2$ loss in our experiments. Table \ref{tab:loss_abl_rpn} shows the results. Thus, for all the results reported in the paper we use $L_2$ loss for feature distillation.

\begin{SCtable}[3][h]
\setlength{\tabcolsep}{4pt}
\scalebox{0.9}{
\begin{tabular}{@{}l|ccc@{}}
\toprule
Methods & $\mathcal{T}_1$ & $\mathcal{T}_2$ & All  \\ \midrule
NST \cite{huang2017like}       & 60.8   & 39.5   & 55.5 \\
AT \cite{zagoruyko2016paying}        & 49.4   & 28.3   & 44.2 \\
$L_2$ Loss      & 64.5   & 40.8   & 58.5 \\ \bottomrule
\end{tabular}}
\caption{\small The effect of using two other Knowledge Distillation techniques, NST \cite{huang2017like} and AT \cite{zagoruyko2016paying}, other than the $L_2$ loss used to distill knowledge in our approach.}
\label{tab:loss_abl_rpn}
\end{SCtable}

\noindent\textbf{Motivation of Using Distillation Loss:} \label{sec:comparison_MT}
When a model is adapted to learn a new task incrementally, the distillation loss -- on the intermediate features generated by the backbone and the output of classification and regression heads -- ensures that they do not deviate much from the model trained on the previous tasks. 
The previous task model acts as a check-point that the incremental model can refer to as a knowledgeable teacher who has expertise in detecting the previously known classes. Once the current set of classes are learned well, the incremental model recursively  serves as the teacher model for the next incremental step. 

The benefit of using distillation loss is also shown in the ablation experiment results showcased in Table~\ref{tab:interplay}, where we selectively turn-on and turn-off distillation loss, keeping all other methodological components unchanged in rows 4 and Table~\ref{tab:interplay}. We see that without using distillation loss, the model incurs a drop of $1.7~\%$ mAP . Gradient preconditioning prepares the model for quick adaptability, ensuring the required plasticity in the network by balancing the intransigence induced by distillation. This is evident from the $4.2~\%$ improvement in mAP while finetuning a distilled model with and without using gradient preconditioning (kindly refer to row 2 and row 6 of Table~\ref{tab:interplay}). 
All these experiments considers the 15+5 configuration on Pascal VOC dataset.

\begin{table}[t]
\centering
\resizebox{0.49\textwidth}{!}{%
\begin{tabular}{@{}c|ccc|ccc|ccc@{}}
\toprule
 & \multicolumn{3}{c|}{$\gamma = 20$} & \multicolumn{3}{c|}{$\gamma = 200$} & \multicolumn{3}{c}{$\gamma = 2000$} \\ \midrule
$\alpha$ & $\mathcal{T}_1$ & $\mathcal{T}_2$ & Both & $\mathcal{T}_1$ & $\mathcal{T}_2$ & Both & $\mathcal{T}_1$ & $\mathcal{T}_2$ & Both \\ \midrule
0.1 & 70.48 & 58.16 & 67.40 & 70.21 & 57.65 & 67.07 & 70.17 & 58.11 & 67.15 \\
0.2 & 70.36 & 57.90 & \textbf{67.76} & 70.52 & 57.99 & 67.39 & 70.67 & 57.55 & 67.39 \\
0.4 & 70.65 & 57.15 & 67.22 & 70.61 & 56.69 & 67.12 & 70.74 & 53.89 & 65.77 \\
0.6 & 71.19 & 54.95 & 67.13 & 70.87 & 54.34 & 66.67 & 70.93 & 52.09 & 64.88 \\
0.8 & 71.98 & 49.21 & 65.71 & 71.58 & 50.78 & 65.63 & 71.34 & 51.11 & 64.77 \\ \bottomrule
\end{tabular}%
}
\vspace{4pt}
\caption{We study the performance of the incremental detector when the warp update interval ($\gamma$; measured as number of iterations) and the weighting factor of the distillation loss ($\alpha$) is changed. All the experiments were performed in the $15+5$ setting. $\mathcal{T}_1$ refers to mAP of the first $15$ classes and $\mathcal{T}_2$ refers to the last $5$ classes from Pascal VOC dataset. `Both Tasks' refers to the mAP of all 20 classes. The best performance is shown in bold.}
\label{tab:alpha-gamma-ablation}
\end{table}

\noindent\textbf{Using a Moving Average Teacher\cite{meanteacherTarvainen}:}
Mean-Teacher \cite{meanteacherTarvainen} proposes to use a moving average teacher to distill the student from. This has found success in semi-supervised learning \cite{luo2018smooth,zhang2019pairwise,meanteacherTarvainen} and domain adaptation  \cite{xu2019self,cai2019exploring,choi2019self,french2017self} settings.
Moving the weights of the teacher network towards the student network is helpful in these settings because the teacher and student are collaboratively learning from a few set of data-points (in semi-supervised setting) or learning in a new domain (in domain adaptation setting where \textit{performance on the earlier domain} can degrade and is not quantified). There is a major difference in the continual learning setting where the student will catastrophically forget the previous task when it learns a new task without access to the previous task data.
We modified our methodology with a moving average teacher and report the results in Table \ref{tab:mean-teacher}.
Updating the teacher with a forgetful student tends to degrade the teachers knowledge on the previous task; and thus is not able to guide the student effectively, as evident from the results. 
The parameters of the teacher $\theta_{t}^{T}$ at time step $t$, are updated with the student's parameters $\theta^{S}$, using the following equation: $\theta_{t}^{T} = \nu \theta_{t}^{T} + (1 - \nu)\theta^{S}$. 
The performance worsens as the teacher model is more aggressively updated with the student's parameters (controlled by $\nu$), across all three settings. 

\begin{table*}[h]
\centering
\resizebox{\textwidth}{!}{%
\begin{tabular}{@{}lllllllllllllllllllllll@{}}
\toprule
 & Setting & aero & cycle & bird & boat & bottle & bus & car & cat & chair & cow & table & dog & horse & bike & person & plant & sheep & sofa & train & tv & mAP \\ \midrule
 & All 20 & 79.4 & 83.3 & 73.2 & 59.4 & 62.6 & 81.7 & 86.6 & 83.0 & 56.4 & 81.6 & 71.9 & 83.0 & 85.4 & 81.5 & 82.7 & 49.4 & 74.4 & 75.1 & 79.6 & 73.6 & 75.2 \\ \midrule
 & Shmelkov \etal \cite{shmelkov2017incremental} & 69.9 & 70.4 & 69.4 & 54.3 & 48.0 & 68.7 & 78.9 & 68.4 & 45.5 & 58.1 & \cellcolor[HTML]{DAE8FC}59.7 & \cellcolor[HTML]{DAE8FC}72.7 & \cellcolor[HTML]{DAE8FC}73.5 & \cellcolor[HTML]{DAE8FC}73.2 & \cellcolor[HTML]{DAE8FC}66.3 & \cellcolor[HTML]{DAE8FC}29.5 & \cellcolor[HTML]{DAE8FC}63.4 & \cellcolor[HTML]{DAE8FC}61.6 & \cellcolor[HTML]{DAE8FC}69.3 & \cellcolor[HTML]{DAE8FC}62.2 & 63.1 \\
 & Ours & 76.0 & 74.6 & 67.5 & 55.9 & 57.6 & 75.1 & 85.4 & 77.0 & 43.7 & 70.8 & \cellcolor[HTML]{DAE8FC}60.1 & \cellcolor[HTML]{DAE8FC}66.4 & \cellcolor[HTML]{DAE8FC}76.0 & \cellcolor[HTML]{DAE8FC}72.6 & \cellcolor[HTML]{DAE8FC}74.6 & \cellcolor[HTML]{DAE8FC}39.7 & \cellcolor[HTML]{DAE8FC}64.0 & \cellcolor[HTML]{DAE8FC}60.2 & \cellcolor[HTML]{DAE8FC}68.5 & \cellcolor[HTML]{DAE8FC}60.5 & \textbf{66.3} \\
 & Ours + MT~\cite{meanteacherTarvainen} ($\nu = 0.99$) & 73.2 & 74.5 & 57.7 & 50.6 & 46.0 & 78.5 & 82.9 & 73.1 & 45.8 & 68.4 & \cellcolor[HTML]{DAE8FC}38.4 & \cellcolor[HTML]{DAE8FC}69.8 & \cellcolor[HTML]{DAE8FC}69.1 & \cellcolor[HTML]{DAE8FC}69.8 & \cellcolor[HTML]{DAE8FC}72.0 & \cellcolor[HTML]{DAE8FC}35.7 & \cellcolor[HTML]{DAE8FC}55.6 & \cellcolor[HTML]{DAE8FC}52.4 & \cellcolor[HTML]{DAE8FC}64.1 & \cellcolor[HTML]{DAE8FC}64.0 & 62.1 \\
 & Ours + MT~\cite{meanteacherTarvainen} ($\nu = 0.9$) & 73.8 & 75.0 & 58.3 & 50.7 & 44.6 & 77.1 & 79.2 & 74.9 & 46.3 & 68.1 & \cellcolor[HTML]{DAE8FC}35.9 & \cellcolor[HTML]{DAE8FC}69.7 & \cellcolor[HTML]{DAE8FC}73.0 & \cellcolor[HTML]{DAE8FC}68.3 & \cellcolor[HTML]{DAE8FC}72.4 & \cellcolor[HTML]{DAE8FC}31.7 & \cellcolor[HTML]{DAE8FC}53.4 & \cellcolor[HTML]{DAE8FC}51.8 & \cellcolor[HTML]{DAE8FC}65.3 & \cellcolor[HTML]{DAE8FC}64.1 & 61.7 \\
\multirow{-5}{*}{{\STAB{\rotatebox[origin=c]{90}{10 + 10}}}} & Ours + MT~\cite{meanteacherTarvainen} ($\nu = 0.8$) & 73.4 & 74.9 & 56.6 & 52.0 & 46.2 & 77.9 & 82.8 & 74.5 & 46.4 & 65.1 & \cellcolor[HTML]{DAE8FC}35.4 & \cellcolor[HTML]{DAE8FC}67.8 & \cellcolor[HTML]{DAE8FC}69.8 & \cellcolor[HTML]{DAE8FC}66.5 & \cellcolor[HTML]{DAE8FC}72.2 & \cellcolor[HTML]{DAE8FC}32.1 & \cellcolor[HTML]{DAE8FC}52.8 & \cellcolor[HTML]{DAE8FC}52.8 & \cellcolor[HTML]{DAE8FC}67.5 & \cellcolor[HTML]{DAE8FC}64.0 & 61.5 \\ \midrule
 & Shmelkov \etal \cite{shmelkov2017incremental} & 70.5 & 79.2 & 68.8 & 59.1 & 53.2 & 75.4 & 79.4 & 78.8 & 46.6 & 59.4 & 59.0 & 75.8 & 71.8 & 78.6 & 69.6 & \cellcolor[HTML]{DAE8FC}33.7 & \cellcolor[HTML]{DAE8FC}61.5 & \cellcolor[HTML]{DAE8FC}63.1 & \cellcolor[HTML]{DAE8FC}71.7 & \cellcolor[HTML]{DAE8FC}62.2 & 65.9 \\
 & Ours & 78.4 & 79.7 & 66.9 & 54.8 & 56.2 & 77.7 & 84.6 & 79.1 & 47.7 & 75.0 & 61.8 & 74.7 & 81.6 & 77.5 & 80.2 & \cellcolor[HTML]{DAE8FC}37.8 & \cellcolor[HTML]{DAE8FC}58.0 & \cellcolor[HTML]{DAE8FC}54.6 & \cellcolor[HTML]{DAE8FC}73.0 & \cellcolor[HTML]{DAE8FC}56.1 & \textbf{67.8} \\
 & Ours + MT~\cite{meanteacherTarvainen} ($\nu = 0.99$) & 72.3 & 76.1 & 65.9 & 53.6 & 55.1 & 74.1 & 82.5 & 75.9 & 46.2 & 72.2 & 64.1 & 74.3 & 81.3 & 75.7 & 76.2 & \cellcolor[HTML]{DAE8FC}42.0 & \cellcolor[HTML]{DAE8FC}53.5 & \cellcolor[HTML]{DAE8FC}54.8 & \cellcolor[HTML]{DAE8FC}71.9 & \cellcolor[HTML]{DAE8FC}67.7 & 66.7 \\
 & Ours + MT~\cite{meanteacherTarvainen} ($\nu = 0.9$) & 66.6 & 76.8 & 66.3 & 50.9 & 52.6 & 74.7 & 82.6 & 75.5 & 45.8 & 69.2 & 59.7 & 73.7 & 81.8 & 74.9 & 76.3 & \cellcolor[HTML]{DAE8FC}40.0 & \cellcolor[HTML]{DAE8FC}50.8 & \cellcolor[HTML]{DAE8FC}52.2 & \cellcolor[HTML]{DAE8FC}70.4 & \cellcolor[HTML]{DAE8FC}65.6 & 65.3 \\
\multirow{-5}{*}{{\STAB{\rotatebox[origin=c]{90}{15 + 5}}}} & Ours + MT~\cite{meanteacherTarvainen} ($\nu = 0.8$) & 71.6 & 76.0 & 64.9 & 51.1 & 52.1 & 73.2 & 78.9 & 74.5 & 45.1 & 68.8 & 61.3 & 75.5 & 76.6 & 68.6 & 76.5 & \cellcolor[HTML]{DAE8FC}39.2 & \cellcolor[HTML]{DAE8FC}53.8 & \cellcolor[HTML]{DAE8FC}49.9 & \cellcolor[HTML]{DAE8FC}71.7 & \cellcolor[HTML]{DAE8FC}65.9 & 64.7 \\ \midrule
 & Shmelkov \etal \cite{shmelkov2017incremental} & 69.4 & 79.3 & 69.5 & 57.4 & 45.4 & 78.4 & 79.1 & 80.5 & 45.7 & 76.3 & 64.8 & 77.2 & 80.8 & 77.5 & 70.1 & 42.3 & 67.5 & 64.4 & 76.7 & \cellcolor[HTML]{DAE8FC}62.7 & 68.3 \\
 & Ours & 78.2 & 77.5 & 69.4 & 55.0 & 56.0 & 78.4 & 84.2 & 79.2 & 46.6 & 79.0 & 63.2 & 78.5 & 82.7 & 79.1 & 79.9 & 44.1 & 73.2 & 66.3 & 76.4 & \cellcolor[HTML]{DAE8FC}57.6 & \textbf{70.2} \\
 & Ours + MT~\cite{meanteacherTarvainen} ($\nu = 0.99$) & 74.1 & 77.1 & 70.4 & 50.6 & 55.3 & 76.2 & 80.2 & 76.8 & 46.8 & 80.0 & 60.0 & 77.5 & 80.9 & 74.6 & 76.1 & 45.8 & 68.9 & 63.0 & 74.4 & \cellcolor[HTML]{DAE8FC}59.3 & 68.4 \\
 & Ours + MT~\cite{meanteacherTarvainen} ($\nu = 0.9$) & 73.2 & 76.2 & 72.0 & 50.2 & 51.2 & 71.2 & 78.2 & 75.3 & 46.9 & 79.3 & 56.9 & 78.8 & 81.1 & 73.7 & 76.6 & 45.4 & 70.1 & 64.3 & 71.4 & \cellcolor[HTML]{DAE8FC}58.8 & 67.5 \\
\multirow{-5}{*}{{\STAB{\rotatebox[origin=c]{90}{15 + 5}}}} & Ours + MT~\cite{meanteacherTarvainen} ($\nu = 0.8$) & 75.3 & 76.9 & 66.0 & 49.0 & 52.7 & 71.4 & 77.8 & 76.3 & 47.7 & 78.7 & 54.8 & 77.0 & 81.6 & 74.2 & 76.3 & 46.5 & 70.0 & 67.0 & 71.2 & \cellcolor[HTML]{DAE8FC}56.3 & 67.3 \\ \bottomrule
\end{tabular}%
}
\vspace{5pt}
\caption{The table reports the results for the case when the teacher model is an exponential moving average of the student model as proposed in Mean-Teacher (MT) \cite{meanteacherTarvainen} in all the three incremental settings (10+10, 15+5 and 19+1) on Pascal VOC dataset. The performances for the incrementally added classes are highlighted with blue background. The best mAP for each setting is shown in {bold}.}
\label{tab:mean-teacher}
\vspace{-15pt}
\end{table*}
\noindent\textbf{Sensitivity of Warp Interval ($\gamma$) and its Relation with Distillation Factor ($\alpha$):}
We experimentally analyze the sensitivity of the proposed methodology to warp update interval ($\gamma$), and its relation with the distillation weighting factor ($\alpha$). We run all the experiments in the $15+5$ setting and report the results in Table~\ref{tab:alpha-gamma-ablation}. 
There is slight degradation in performance when we update the warp layers less frequently, keeping the distillation weighting factor constant (analyze the table row-wise). This is because the warp layers that are updated more frequently tend to precondition the gradients more effectively. 

When we increase the distillation weighting factor, the network gets more adept in retaining past knowledge restricting it to learn the new classes effectively. This characteristic is consistent across  $\gamma = 20 / 200 / 2000$ iterations, where accuracy of $\mathcal{T}_1$ goes up and that of $\mathcal{T}_2$ goes down. We observe the same trend in Table 7 of the main paper which had used only distillation to address forgetting, while we use distillation, gradient preconditioning and finetuning in the current experiment.


\begin{table*}
\parbox{1\linewidth}{
\centering
\resizebox{\textwidth}{!}{%
\begin{tabular}{@{}l|aaaaaaaaaabbbbbbbbbb|l@{}}
\toprule
Train with & aero        & cycle         & bird          & boat          & bottle        & bus           & car           & cat         & chair         & cow           & table         & dog           & horse       & bike          & person        & plant         & sheep       & sofa          & train         & tv            & mAP           \\ \midrule
All 20     & 79.4        & 83.3          & 73.2          & 59.4          & 62.6          & 81.7          & 86.6          & 83.0          & 56.4          & 81.6          & 71.9          & 83.0            & 85.4        & 81.5          & 82.7          & 49.4          & 74.4        & 75.1          & 79.6          & 73.6          & 75.2          \\
First 10   & 78.6        & 78.6          & 72.0           & 54.5          & 63.9          & 81.5          & 87.0            & 78.2        & 55.3          & 84.4          & -             & -             & -           & -             & -             & -             & -           & -             & -             & -             & 73.4          \\
Std Training    & 35.7 & 9.1   & 16.6 & 7.3  & 9.1    & 18.2 & 9.1  & 26.4 & 9.1   & 6.1  & 57.6  & 57.1 & 72.6  & 67.5 & 73.9   & 33.5  & 53.4  & 61.1 & 66.5  & 57.0   & 37.3 \\ \midrule
Shmelkov \etal \cite{shmelkov2017incremental}  & 69.9        & 70.4          & {69.4} & 54.3          & 48.0            & 68.7          & 78.9          & 68.4        & {45.5} & 58.1          & 59.7          & {72.7} & 73.5        & {73.2} & 66.3          & 29.5          & 63.4        & {61.6} & {69.3} & {62.2} & 63.1          \\
Faster ILOD \cite{peng2020faster} & 72.8 & 75.7 & 71.2 & 60.5 & 61.7 & 70.4 & 83.3 & 76.6 & 53.1 & 72.3 & 36.7 & 70.9 & 66.8 & 67.6 & 66.1 & 24.7 & 63.1 & 48.1 & 57.1 & 43.6 & 62.2 \\
ORE~\cite{joseph2021open} & 63.5 & 70.9 & 58.9 & 42.9 & 34.1 & 76.2 & 80.7 & 76.3 & 34.1 & 66.1 & 56.1 & 70.4 & 80.2 & 72.3 & 81.8 & 42.7 & 71.6 & 68.1 & 77.0 & 67.7 & 64.6 \\
DMC* \cite{zhang2020class} & 68.6 & 71.2 & 73.1 & 48.1 & 56.0 & 64.4 & 81.9 & 77.8 & 49.4 & 67.8 & 61.5 & 67.7 & 67.5 & 52.2 & 74.0 & 37.8 & 63.0 & 55.5 & 65.3 & 72.4 & 63.8 \\
DMC \cite{zhang2020class} & 73.9 & 81.7 & 72.7 & 54.6 & 59.2 & 73.3 & 85.2 & 83.3 & 52.9 & 68.1 & 62.6 & 75.0 & 69.0 & 63.4 & 80.3 & 42.4 & 60.3 & 61.5 & 72.6 & 74.5 & 68.3 \\ \midrule
Ours       & {76.0} & {74.6} & 67.5          & {55.9} & {57.6} & {75.1} & {85.4} & {77.0} & 43.7          & {70.8} & {60.1} & 66.4          & {76.0} & 72.6          & {74.6} & {39.7} & {64.0} & 60.2          & 68.5          & 60.5          & {66.3} \\
Ours + aux data & 75.5 & 81.3 & 69.2 & 54.8 & 57.2 & 77.3 & 84.2 & 79.3 & 47.4 & 70.5 & 63.1 & 78.2 & 77.5 & 70.9 & 80.7 & 43.3 & 63.1 & 66.5 & 74.6 & 70.7 & \textbf{69.3} \\
\bottomrule
\end{tabular}%
}
\vspace{4pt}
\caption{ Per-class AP and overall mAP on PASCAL VOC when 10 new classes (\emph{right}) are added to an object detector trained initially on first 10 classes (\emph{left}). DMC*~\cite{zhang2020class} refers to a variant of DMC~\cite{zhang2020class} which uses auxiliary data that does not include Pascal VOC classes.
}
\label{tab:10_plus_10}
}
\hfill
\parbox{1\linewidth}{
\centering
\resizebox{\textwidth}{!}{%
\begin{tabular}{@{}l|aaaaaaaaaaaaaaaaaaab|l@{}}
\toprule
Train with & aero          & cycle         & bird          & boat          & bottle      & bus           & car           & cat           & chair         & cow         & table         & dog           & horse         & bike          & person        & plant         & sheep         & sofa          & train         & tv            & mAP           \\ \midrule
All 20     & 79.4          & 83.3          & 73.2          & 59.4          & 62.6        & 81.7          & 86.6          & 83.0            & 56.4          & 81.6        & 71.9          & 83.0            & 85.4          & 81.5          & 82.7          & 49.4          & 74.4          & 75.1          & 79.6          & 73.6          & 75.2          \\
First 19   & 76.3          & 77.3          & 68.4          & 55.4          & 59.7        & 81.4          & 85.3          & 80.3          & 47.8          & 78.1        & 65.7          & 77.5          & 83.5          & 76.2          & 77.2          & 46.6          & 71.4          & 65.8          & 76.5          & -             & 67.5          \\
Std Training    & 16.6          & 9.1           & 9.1           & 9.1           & 9.1         & 8.3           & 35.3          & 9.1           & 0.0             & 22.3        & 9.1           & 9.1           & 9.1           & 13.7          & 9.1           & 9.1           & 23.1          & 9.1           & 15.4          & 50.7          & 14.3          \\ \midrule
Shmelkov \etal \cite{shmelkov2017incremental}   & 69.4          & {79.3} & {69.5} & {57.4} & 45.4        & 78.4          & 79.1          & {80.5} & 45.7          & 76.3        & {64.8} & 77.2          & 80.8          & 77.5          & 70.1          & 42.3          & 67.5          & 64.4          & {76.7} & {62.7} & 68.3          \\
Faster ILOD \cite{peng2020faster} & 64.2 & 74.7 & 73.2 & 55.5 & 53.7 & 70.8 & 82.9 & 82.6 & 51.6 & 79.7 & 58.7 & 78.8 & 81.8 & 75.3 & 77.4 & 43.1 & 73.8 & 61.7 & 69.8 & 61.1 & 68.6 \\
ORE \cite{joseph2021open} & 67.3 & 76.8 & 60.0 & 48.4 & 58.8 & 81.1 & 86.5 & 75.8 & 41.5 & 79.6 & 54.6 & 72.8 & 85.9 & 81.7 & 82.4 & 44.8 & 75.8 & 68.2 & 75.7 & 60.1 & 68.9 \\
DMC* \cite{zhang2020class} & 65.3 & 65.8 & 73.2 & 43.8 & 57.1 & 73.3 & 83.1 & 79.3 & 45.4 & 74.3 & 55.1 & 82.0 & 68.7 & 62.6 & 74.9 & 42.3 & 65.2 & 67.5 & 67.8 & 64.0 & 65.5 \\
DMC \cite{zhang2020class} & 75.4 & 77.4 & 76.4 & 52.6 & 65.5 & 76.7 & 85.9 & 80.5 & 51.2 & 76.1 & 63.1 & 83.3 & 74.6 & 73.7 & 80.1 & 44.6 & 67.5 & 68.1 & 74.4 & 69.0 & 70.8 \\ \midrule
Ours       & {78.2} & 77.5          & 69.4          & 55.0            & {56.0} & {78.4} & {84.2} & 79.2          & {46.6} & {79.0} & 63.2          & {78.5} & {82.7} & {79.1} & {79.9} & {44.1} & {73.2} & {66.3} & 76.4          & 57.6          & {70.2} \\
Ours + aux data & 79.6 & 81.3 & 72.2 & 55.3 & 60.2 & 77.4 & 84.4 & 77.5 & 48.5 & 79.2 & 63.3 & 80.3 & 82.2 & 77.2 & 80.8 & 45.7 & 72.2 & 67.8 & 73.3 & 63.7 & \textbf{71.1} \\
\bottomrule
\end{tabular}%
}\vspace{4pt}
\caption{ Per-class AP and overall mAP when the tvmonitor class from PASCAL VOC is added to the detector, trained on all other classes. DMC* \cite{zhang2020class} refers to a variant of DMC \cite{zhang2020class} which uses auxiliary data that does not include Pascal VOC classes.}
\label{tab:19_plus_1}
\vspace{-17pt}
}
\end{table*}
\noindent\textbf{Comparison with DMC~\cite{zhang2020class}:}
In the process of incrementally learning a new set of classes, Deep Model Consolidation (DMC) \cite{zhang2020class} trains a new object detector just for these classes as the first step. Next, a third object detector is trained by consolidating the base detector (trained on initial classes) and the model trained on new classes using unlabelled auxiliary data via distillation loss. In the experimental analysis on Pascal VOC dataset, $98{,}495$ images from Microsoft COCO are used as auxiliary data. Our proposed methodology requires only two models (base and incremental) to work and does not make use of any auxiliary data. For a fair comparison with DMC \cite{zhang2020class}, we also modify our methodology to make use of auxiliary data (without using any supervised labels) in a simple and naive way. We use the auxiliary data to define an additional constraint in the distillation loss as follows:  
\vspace{-1pt}
\begin{align}
        \mathcal{L}_{Distill\_Aux\_Data} 
        =~ & \mathcal{L}_{Distill} + \sum_{\bm{X}\in Aux\_Data}  \mathcal{L}_{Reg}(\bm{F}_t, \bm{F}_{t-1}); \notag \\
        &~ s.t., \bm{F}_t = \mathcal{F}_{Backbone}^{ t}(\bm{X})
    \label{eqn:warp_loss}
\end{align}

\noindent where $\mathcal{L}_{Reg}$ is the $L_2$ regression loss between features ($\bm{F}_t$) from the backbone of the current model ($\mathcal{F}_{Backbone}^{t}$) and the backbone features ($\bm{F}_{t-1}$) from the model trained on all previous classes. $\mathcal{L}_{Distill}$ is the distillation loss defined in Equation 3 of the main paper. 
This additional constraint will ensure that the backbone features of the incremental model do not deviate much from the base model even for images belonging to the auxiliary data, thus providing some additional supervisory cue. As shown in Tables~\ref{tab:10_plus_10} and~\ref{tab:19_plus_1}, we see that using the auxiliary data further improves the performance of our proposed methodology. We compare against DMC \cite{zhang2020class} (which uses all MS COCO data as auxiliary data) and DMC* \cite{zhang2020class} (where auxiliary data excludes Pascal VOC class instances from MS COCO dataset). We report the results for 10+10 and 19+1 setting only as 15+5 results are not available for DMC \cite{zhang2020class}.

\noindent\textbf{Time and Memory Requirements}
There are two key components in our proposed methodology: (1) meta-learning the gradient preconditioning matrix; and (2) distillation from the model trained on the previous task. In Table~\ref{tab:expense}, we systematically analyze the training time and GPU memory requirement for each of these phases.
\begin{table}[H]
\centering
\resizebox{0.49\textwidth}{!}{%
\begin{tabular}{@{}cc|ll@{}}
\toprule
Distillation & Gradient Preconditioning & Time (sec / iter) & GPU Memory (MB) \\ \midrule
$\times$ & $\times$ & 0.45 & 2598 \\
\checkmark & $\times$ & $0.66_{~(+ 0.21)}$ & $3626_{~(+ 1028)}$ \\
\checkmark & \checkmark & $1.07_{~(+ 0.41)}$ & $4921_{~(+ 1295)}$ \\ \bottomrule
\end{tabular}%
}
\vspace{3pt}
\caption{Time and GPU memory expense for our proposed methodology. 
The subscript indicates the additional expense incurred due to the corresponding component.
}
\label{tab:expense}
\vspace{-10pt}
\end{table}
Once a model is trained using distillation and gradient preconditioning, it is finetuned in the next stage, which takes 0.44 sec/iter. Typically, we train the model with distillation and gradient preconditioning for 18000 iterations in $10+10$ setting and then finetune for 3000 iterations. Hence the total compute time incurred is $5.71$ hours $([1.07 \times 18000] + [0.44 \times 3000])$. 
Experiments in $15+5$ and $19+1$  setting is run for 12000 and 8000 iterations, incurring $3.93$ and $2.74$ hours respectively.
All the timings were measured on a single server equipped with four NVIDIA GTX 1080-Ti GPUs.


\vspace{-9pt}
\section{Conclusion} \label{sec:conclusion}
The existing incremental object detection approaches are based on knowledge distillation, which helps in retaining old learning at the cost of a reduced adaptability to new tasks. 
In this work, we propose a meta-learning approach to object detection, that learns to precondition the gradient updates such that information across incremental tasks is automatically shared. This helps the model not only to retain old knowledge but also to adapt flexibly to new tasks. 
The meta-learned incremental object detector outperforms the current best methods on two benchmark datasets. 
Further, our extensive ablation experiments brings out the contributions of each constituent components of the methodology.
Extending our methodology to single stage detectors \cite{nie2019enriched,wang2019learning,pang2019efficient} and incremental versions of related problem settings such as action recognition \cite{khan2015recognizing}, object counting \cite{cao2020d2det} and instance segmentation \cite{cholakkal2019object} are interesting and important research directions. 


\section*{Acknowledgements}
This work has been partly supported by the funding received from DST through the IMPRINT program (IMP/2019/000250), the TCS PhD Fellowship and VR starting grant (2016-05543).
\ifCLASSOPTIONcaptionsoff
  \newpage
\fi

\bibliographystyle{IEEEtran}
\bibliography{egbib}

\vspace{-30pt}
\begin{IEEEbiography}
[
{\includegraphics[width=1in,height=1.5in,clip,keepaspectratio]{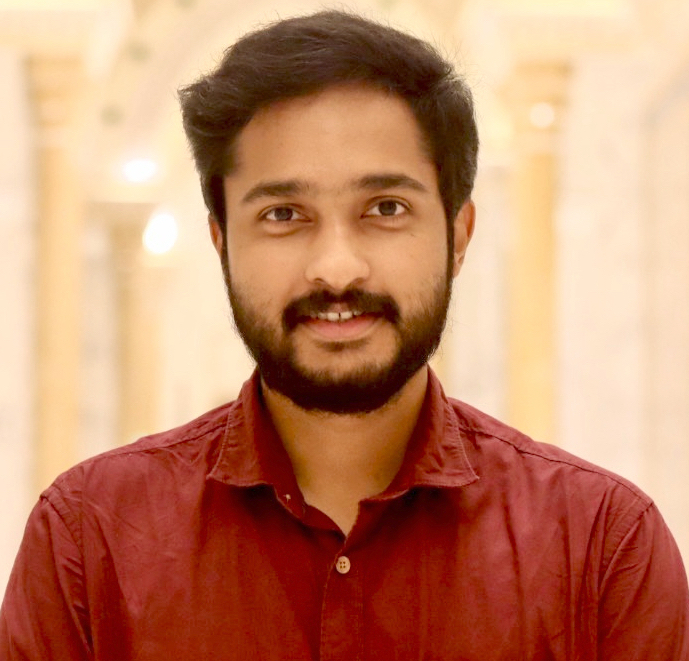}}
]
{Joseph K J} is a PhD student at Indian Institute of Technology Hyderabad, India. His research works have been published at top venues like TPAMI, NeurIPS, CVPR, IJCAI and BMVC. He received the Excellence in Research Award from IIT Hyderabad in 2020 and is supported by the TCS PhD Fellowship. He has been a research intern at Inception Institute of AI, MBZ University of AI, University of Tokyo and Google Research. He actively works in developing machine learning models for computer vision.
\end{IEEEbiography}

\begin{IEEEbiography}
[
{\includegraphics[width=1in,height=1.5in,clip,keepaspectratio]{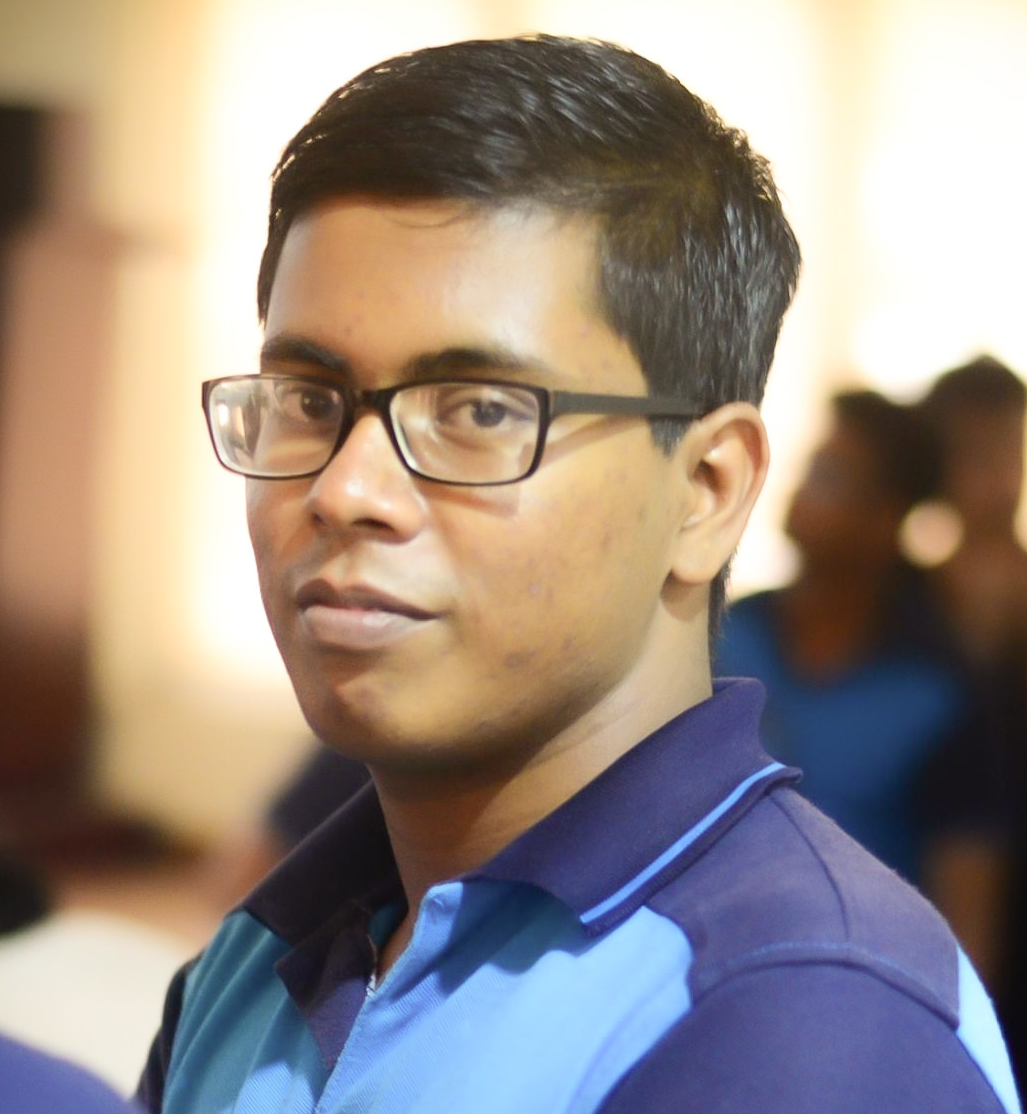}}
]
{Jathushan Rajasegaran} is currently pursuing PhD at University of California, Berkeley. Previously, he worked at the Inception Institute of Artificial Intelligence (IIAI), UAE as a Research Associate. He completed his BS from University of Moratuwa, Sri Lanka and was awarded Gold Medal for his excellent performance. He has published his research in top venues such as CVPR, WWW and NeurIPS. His research interest lies in computer vision, deep learning and tracking.
\end{IEEEbiography}

\vspace{-110pt}
\begin{IEEEbiography}
[
{\includegraphics[width=1in,height=1.5in,clip,keepaspectratio]{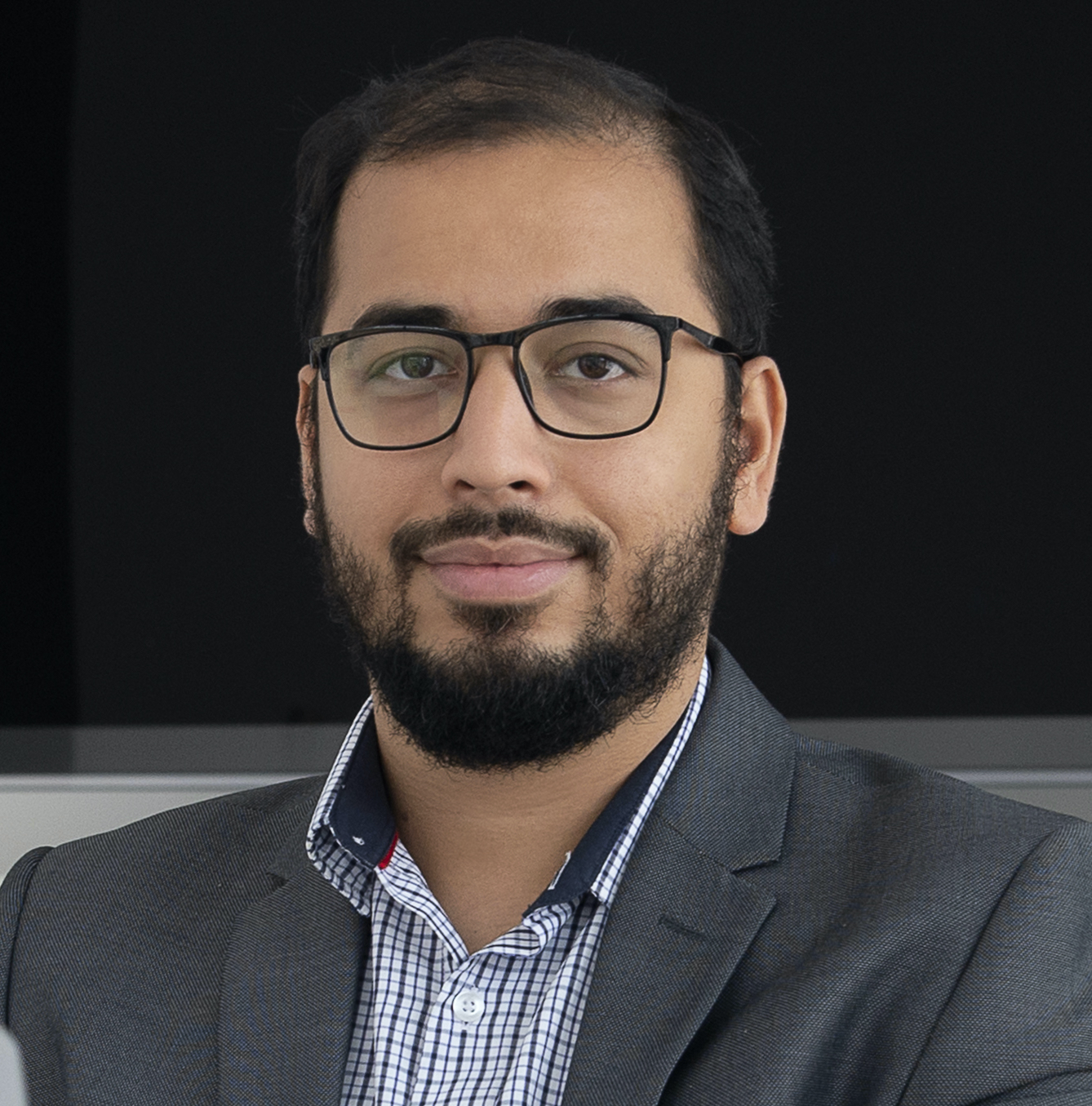}}
]
{Salman Khan} is an Associate Professor at MBZ University of Artificial Intelligence. He has been an Adjunct faculty member with Australian National University since 2016. He worked as a Senior scientist with the Inception Institute of AI (2018-2020) and as a Research Scientist with Data61-CSIRO from 2016-2018. He has been awarded the outstanding reviewer award at IEEE CVPR multiple times, won the best paper award at 9th ICPRAM 2020, and won 2nd prize in the NTIRE Image Enhancement Competition alongside CVPR 2019. He served as a program committee member for several premier conferences including CVPR, ICCV, ICLR, ECCV and NeurIPS. He received his Ph.D. degree from The University of Western Australia in 2016. His thesis received an honorable mention on the Dean’s List Award. He has published over 80 papers in top scientific journals and conferences. His research interests include computer vision and machine learning.
\end{IEEEbiography}

\vspace{-110pt}
\begin{IEEEbiography}
[
{\includegraphics[width=1in,height=1.5in,clip,keepaspectratio]{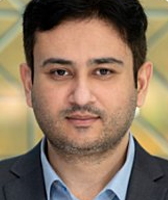}}
]
{Fahad Khan} is a faculty member at MBZUAI, UAE and Linköping University, Sweden. From 2018 to 2020 he worked as a Lead Scientist at the Inception Institute of Artificial Intelligence (IIAI), Abu Dhabi, UAE. He received the M.Sc. degree in Intelligent Systems Design from Chalmers University of Technology, Sweden and a Ph.D. degree in Computer Vision from Autonomous University of Barcelona, Spain. He has achieved top ranks on various international challenges (Visual Object Tracking VOT: 1st 2014 and 2018, 2nd 2015, 1st 2016; VOT-TIR: 1st 2015 and 2016; OpenCV Tracking: 1st 2015; 1st PASCAL VOC 2010). His research interests include a wide range of topics within computer vision and machine learning, such as object recognition, object detection, action recognition and visual tracking. He has published articles in high-impact computer vision journals and conferences in these areas. He serves as a regular program committee member for leading computer vision and artificial intelligence conferences such as CVPR, ICCV, and ECCV.
\end{IEEEbiography}

\vspace{-110pt}
\begin{IEEEbiography}
[
{\includegraphics[width=1in,height=1.5in,clip,keepaspectratio]{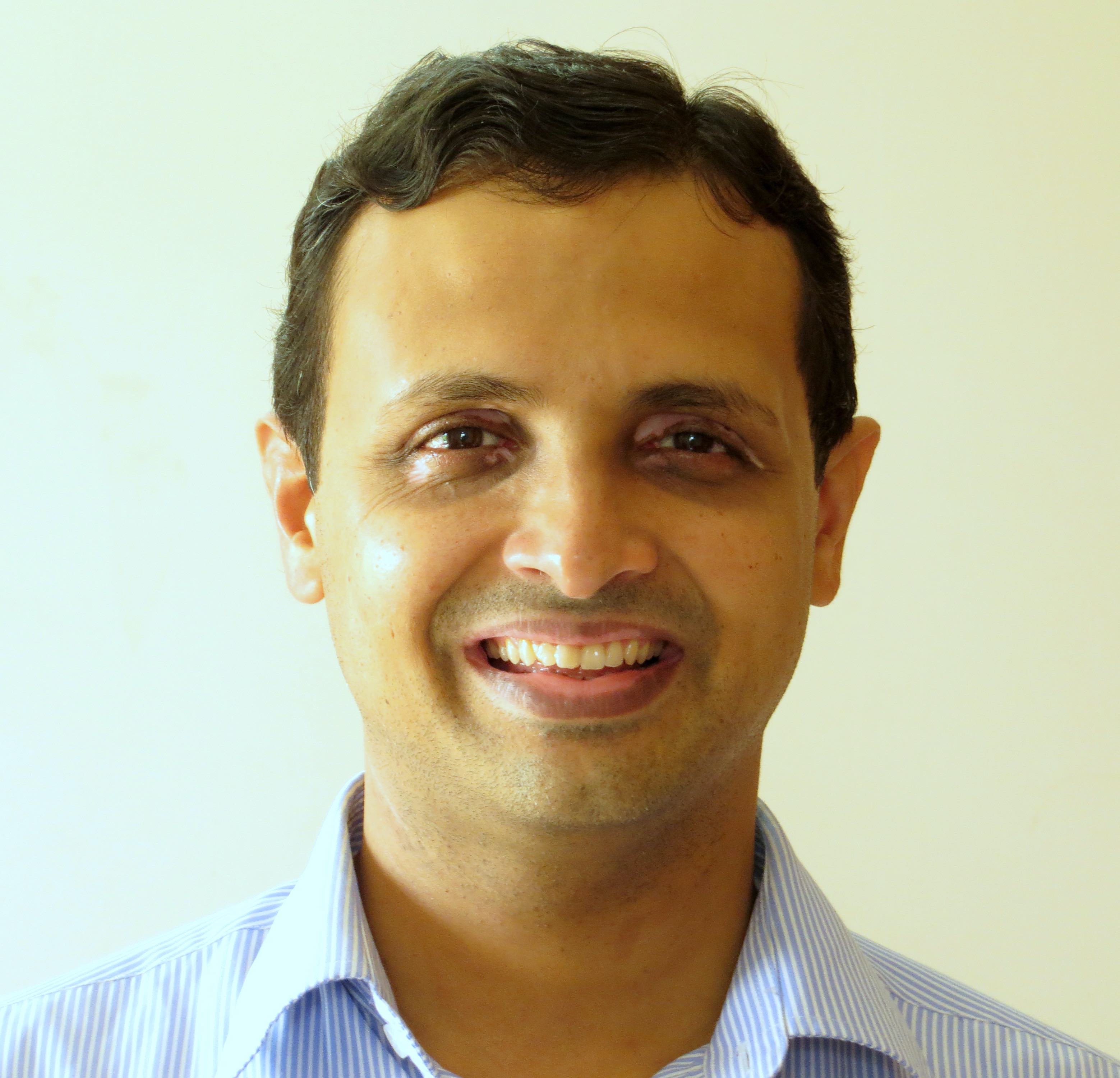}}
]
{Vineeth N Balasubramanian} is an Associate Professor in the Department of Computer Science and Engineering at the Indian Institute of Technology, Hyderabad (IIT-H). His research interests include deep learning, machine learning, and computer vision, with a focus on explainable deep learning and learning with limited supervision. His research has resulted in over 100 peer-reviewed publications at various international venues, including top-tier venues such as ICML, CVPR, NeurIPS, ICCV, KDD, ICDM, and IEEE TPAMI. He regularly serves as a Senior PC/Area Chair for conferences such as CVPR, ICCV, AAAI, and IJCAI, with recent awards as Outstanding Reviewer at ICLR 2021, CVPR 2019, ECCV 2020, etc. He is also a recipient of the Google Research Scholar Award in 2020-21. For more details, please see \url{https://iith.ac.in/~vineethnb/}.
\end{IEEEbiography}









\end{document}